\pdfoutput=1
\documentclass[review]{elsarticle}
\usepackage{subfigure}
\usepackage{longtable}
\usepackage{algorithm, algorithmic}
\usepackage{bm}
\usepackage{booktabs}
\usepackage{multirow}
\usepackage{geometry}
\usepackage{amsfonts,amssymb}
\usepackage{amsmath}
\usepackage{array}

\usepackage{lineno,hyperref}
\modulolinenumbers[5]

\journal{arXiv}

\begin{document}

\begin{frontmatter}

\title{Online Continual Learning via the Meta-learning Update with Multi-scale Knowledge Distillation and Data Augmentation}


\author[mymainaddress]{Ya-nan Han}

\author[mymainaddress]{Jian-wei Liu\corref{mycorrespondingauthor}}
\cortext[mycorrespondingauthor]{Corresponding author}
\ead{liujw@cup.edu.cn}

\address[mymainaddress]{Department of Automation, College of Information Science and Engineering, China University of Petroleum Beijing (CUP)
}

\begin{abstract}
Continual learning aims to rapidly and continually learn the current task from a sequence of tasks, using the knowledge obtained in the past, while performing well on prior tasks. A key challenge in this setting is the stability-plasticity dilemma existing in current and previous tasks, i.e., a high-stability network is weak to learn new knowledge in an effort to maintain previous knowledge. Correspondingly, a high-plasticity network can easily forget old tasks while dealing with well on the new task. Compared to other kinds of methods, the methods based on experience replay have shown great advantages to overcome catastrophic forgetting. One common limitation of this method is the data imbalance between the previous and current tasks, which would further aggravate forgetting. Moreover, how to effectively address the stability-plasticity dilemma in this setting is also an urgent problem to be solved. In this paper, we overcome these challenges by proposing a novel framework called Meta-learning update via Multi-scale Knowledge Distillation and Data Augmentation (MMKDDA). Specifically, we apply multi-scale knowledge distillation to grasp the evolution of long-range and short-range spatial relationships at different feature levels to alleviate the problem of data imbalance. Besides, our method mixes the samples from the episodic memory and current task in the online continual training procedure, thus alleviating the side influence due to the change of probability distribution. Moreover, we optimize our model via the meta-learning update resorting to the number of tasks seen previously, which is helpful to keep a better balance between stability and plasticity. Finally, our experimental evaluation on four benchmark datasets shows the effectiveness of the proposed MMKDDA framework against other popular baselines, and ablation studies are also conducted to further analyze the role of each component in our framework.
\end{abstract}


\end{frontmatter}

\section{Introduction}

Deep neural networks perform well on various computer vision problems
\cite{1girshick2015fast,2krizhevsky2012imagenet,3long2015fully,4zhang2018fully}, such as semantic segmentation
\cite{5pereira2019adaptive}, image classification
\cite{6chandra2020ae}, object recognition
\cite{7alom2021inception}. However, catastrophic forgetting problem
\cite{8french1993catastrophic,9goodfellow2013empirical,10kirkpatrick2017overcoming} arises when the agent is required to learn new tasks while maintaining the ability to perform well on the old tasks. This is due to two key factors: (i) the updates for new tasks can remove the learned old knowledge obtained from the old tasks
\cite{10kirkpatrick2017overcoming,11shin2017continual,12kemker2018measuring}; (ii) In terms of this learning scenario, when the new task arrives, replaying all training data from previous tasks is prohibitive for the agent. 

Continual learning
\cite{13ring1995continual,14thrun1998lifelong,15thrun1995lifelong} aims to tackle these issues, and requires an agent to learn continuously from a stream of tasks by leveraging the knowledge obtained in the past(plasticity) while remembering how to perform previously learned tasks(stability)
\cite{16mermillod2013stability}. Up till now the leading works about continual learning
\cite{17robins1993catastrophic,18maltoni2019continuous,19parisi2019continual} focus on studying how to effectively learn new knowledge while not forgetting the previous task. Different methods have been presented to overcome these challenges, such as regularization-based approaches
\cite{10kirkpatrick2017overcoming},\cite{20li2016learning,21pomponi2020efficient,22schwarz2018progress}, architecture-based approaches
\cite{23yoon2018lifelong,24ebrahimi2020adversarial}, and memory-based approaches
\cite{25chaudhry2019efficient,26chaudhry2019continual,27riemer2018learning}. It is noted that although initial successes have been achieved, there are still a lot of open challenges. Regularization-based approaches extend loss function with a term to mitigate catastrophic forgetting. The core idea of these methods is to control the updates of the model parameters, and thereby consolidate prior knowledge when a new task comes. However, it is worth noting that when the number of tasks is large, and after the models are trained many times, most model parameters prefer not to change because of the regularization constraint. Therefore, it is difficult to perform larger updates for learning the model parameters of new knowledge. In contrast, architecture-based approaches train separate models for each task, and then a selection criterion is set to decide which model should be applied during the inference stage. Obviously, these types of approaches are poorly scalable for a large number of tasks due to the ever-increasing architectures as the number of tasks increases.

In addition, in recent studies, existing works mainly focus on the situations where the whole task’s training data comes at each step and allow the agent to learn the current tasks for many epochs. However, this assumption is not suitable for many real-world scenarios where the training data is observed sequentially in a stream form. Based on this fact, in this paper, we consider a more realistic but difficult scenario, called online continual learning. In the online continual learning setting, to alleviate catastrophic forgetting, the feasible approaches usually require storing a small portion of previous training data. In this case, compared to storing the whole old data for replay, there is a limited memory budget that is used to perform experience replay. However, a cumbersome data imbalance problem between previous and current tasks can be caused due to the memory constraint, which can indirectly lead to the inherent problem in the field of continual learning, i.e., the stability-plasticity dilemma
\cite{16mermillod2013stability}. Moreover, how to effectively adapt to the new task is also a key issue. In particular, higher plasticity can cause the forgetting of previous tasks while learning well on new tasks. However, higher stability can weaken the ability to perform new tasks due to the effort to preserve old knowledge. It is at odds with each other. Based on the above discussion, a favorable model should be quickly adapted to new tasks while maintaining the stability of the model. Furthermore, it should also make a balance between plasticity and stability. Some related works either are devoted to improving the plasticity of the network
\cite{28finn2019online} or focusing on catastrophic forgetting
\cite{25chaudhry2019efficient,26chaudhry2019continual}. So how to effectively overcome the stability-plasticity dilemma is a key problem in this setting.

Inspired by the above issues, in this paper, to tackle the stability-plasticity dilemma, we propose a novel framework called Meta-learning via Multi-scale Knowledge Distillation and Data Augmentation (MMKDDA) specifically designed for online continual learning setting. More concretely, to avoid catastrophic forgetting, MMKDDA keeps an episodic memory for prior tasks which is applied to jointly train with data of the current task. Different from the conventional experience replay used in the literature
\cite{26chaudhry2019continual}, to reduce the bias given rise to the data imbalance, MMKDDA employs a multi-scale knowledge distillation regularizer to constrain the evolution of the representation at different feature levels. Meanwhile, we consider that when the memory replay is performed, the marginal distributions are different between the samples of a new task and a limited number of ones obtained from episodic memory. So, in order to quickly adapt to the desired task, an effective data augmentation strategy, called Cutmix
\cite{29yun2019cutmix} augmentation, is adopted to generate new augmented data. In this work, we attempt to adopt Cutmix to mix the image-label of multiple data samples randomly where the samples are obtained from the current task and previous tasks respectively. The probability distribution of generated examples is interpolated between the probability distribution of the old and current tasks, which can further mitigate the side influence due to the change of the data distribution over the tasks. Furthermore, during training, MMKDDA devises a meta-update rule based on the inherent relationships between the sequential tasks to further keep an equilibrium between stability and plasticity as the number of tasks adds.

In summary, the major contributions of this work are as follows:

1. We utilize the data augmentation strategy to the mini-batch examples of the current task and additionally a batch of the memory replay examples, and boost the capacity of MMKDDA to adapt the change of data probability distribution.

2. Furthermore, consider that previous task data in the episodic memory are limited compared to the current training data, which causes a bias towards the current task. A multi-scale knowledge distillation strategy is used to alleviate this problem.

3. We introduce a monotonic scheduling strategy based on the number of tasks seen previously to fine-tune the meta-learning rate. This updated strategy would be beneficial to learn the inherent relationships between consecutive tasks.

4.We conduct comprehensive experiments to demonstrate our improvements against a series of state-of-the-art approaches, and experiments on several commonly used datasets show that our method achieves state-of-the-art performance.

The rest of this paper is structured as follows. In section 2, the related work is introduced. In section 3, we illustrate the problem of continual learning and then present our MMKDDA which can learn continuous tasks in the online setting while maintaining the balance of stability and plasticity. In section 4, contrast experiments and ablate studies are performed to demonstrate our improvements against other baselines and the contribution of different factors in our experiments. Finally, in section 5, we conclude our whole paper and discuss future work.

\section{Related Work}

The ultimate goal of continual learning is to achieve good performance on a new task while not forgetting how to perform the old tasks. Continual learning called lifelong learning
\cite{14thrun1998lifelong,30thrun1995lifelong} or never-ending learning date back to the 1990s. However, in recent years with the development of deep learning, the decades-old problem of continual learning has captured a lot of attention from the community again. Related works mainly focus on the problem of forgetting when the model is trained on sequential tasks. The existing continual learning approaches to alleviate catastrophic forgetting can be coarsely grouped into three categories: regularization strategy, architecture strategy, and memory strategy.

\textbf{Regularization-based methods.} The methods of this category try to add a special regularization term to constrain the updating of the model parameters to overcome catastrophic forgetting. One representative approach, elastic weight consolidation(EWC)
\cite{10kirkpatrick2017overcoming}, applies the Fisher information matrix(FIM) to measure the importance of weights. By this means, EWC can control the changing of weights when the new task comes, that means, there is a smaller change for important weights and vice versa. Based on this, Amer et al.
\cite{31amer2019reducing} present a new approach composed of dynamic information balancing and EWC regularization. Zenke et al.
\cite{32zenke2017continual} present Synaptic Intelligence(SI) approach, SI can compute the importance of weight by considering the cumulative change of distance after learning new tasks. Obviously, SI is more intuitive and persuasive compared with EWC in terms of the approximation about the importance of weights. Chaudhry et al.
\cite{33chaudhry2018riemannian} study the problem of forgetting and intransigence in continual learning and then propose the RWalk method. Other relevant works are the approaches based on knowledge distillation
\cite{34hinton2015distilling}. Li et al.
\cite{20li2016learning} present Learning without Forgetting(LwF) which learns a separate classifier for every task. When the new task is observed, LwF can apply knowledge distillation to constraint the change of parameters by combining the outputs obtained by the old and current classifiers respectively. Schwarz et al. 
\cite{22schwarz2018progress}introduce a conceptually simple and scalable framework called Progress \&Compress which divides the learning into two stages: progression and compression. In the progress stage, the model focuses on positive transfer, that means, the model learns new task via the knowledge obtained from the previous tasks. In the compress stage, the newly learned knowledge is integrated into the old model via knowledge distillation to alleviate forgetting. Based on this, a series of methods
\cite{35hou2018lifelong,36kim2019incremental,37zhang2020class,38zhao2020maintaining} have been proposed. It is worth noting that these methods based on regularization strategy can mitigate forgetting without expanding the architecture of the network or storing partial data for previous tasks. However, these kinds of methods have the limited ability to overcome catastrophic forgetting, especially in the online learning setting, and ignore the problem of plasticity in the learning process.

\textbf{Architecture-based methods.} These approaches attempt to train a separate model for each incoming task to alleviate catastrophic forgetting. Rusu et al.
\cite{39rusu2016progressive} present a progressive neural network(PNN) model. PNN combats the forgetting of old knowledge by fixing the parameters of the network learned for prior tasks and learns the new knowledge when a new task arrives by expanding the network. However, the number of parameters would gradually increase when the number of tasks rises. Roy et al.
\cite{40roy2020tree} propose a hierarchical network based on a tree structure for continual learning. However, the training algorithm is not efficient and this method also requires more space. Consider the overlapping knowledge among the different tasks, Yoon et al.
\cite{23yoon2018lifelong} propose a method called dynamic expansion network(DEN). This method can adaptively decide the network structure based on the correlation of knowledge and then improve the efficiency of the network capacity. Other relevant works such as ExpertGate
\cite{41aljundi2017expert} and ACL
\cite{24ebrahimi2020adversarial} enjoy similar advantages as DEN. Although the methods of this category can overcome catastrophic forgetting to a certain extent, it is poorly suitable for large tasks because the network architecture will gradually increase as the number of task increase
\cite{19parisi2019continual,42delange2021continual}.

Memory-based methods. Memory-based methods follow a relatively simple idea to mitigate catastrophic forgetting. Early studies
\cite{43robins1995catastrophic} find that the problem of catastrophic forgetting can be reduced by rehearsal or pseudo-rehearsal. As the era of deep learning comes, this memory strategy has seen a surge of interest from the community again. Consider that it is also important to improve the ability of the learner to transfer knowledge in the field of continual learning. Lopez-Paz et al.
\cite{44lopez2017gradient} present the Gradient Episodic Memory(GEM), which employs an inequality constraint using episodic memory to avoid forgetting. To further improve the training efficiency, Chaudhry et al.
\cite{25chaudhry2019efficient} propose AGEM based on GEM. Pham et al.
\cite{45pham2020bilevel} aim to improve the generalization and the learning of the current tasks using the knowledge obtained on previous tasks, and then propose the BCL approach based on bilevel learning
\cite{46colson2007overview} and episodic memory. Buzzega et al.
\cite{47buzzega2020dark} propose to alleviate catastrophic forgetting by mixing rehearsal with knowledge distillation and regularization. Other relevant works
\cite{26chaudhry2019continual,27riemer2018learning} mainly focus on joint train combining the current training data and the episodic memory. Along with the development of the generative model, some approaches based on auto-encoder
\cite{48bengio2013representation}, VAE
\cite{49kingma2014auto}, or GAN
\cite{50goodfellow2020generative}, are proposed, such as CLDGR
\cite{11shin2017continual}, BIC
\cite{51wu2019large}, and DGM
\cite{52ostapenko2019learning}. Compared with the regularization strategy, these strategies are regarded as an effective strategy to overcome catastrophic forgetting. However, these methods also have their limitations, such as requiring some additional memory space.

\textbf{Data Augmentation.} Recent works demonstrate that data augmentation can improve the training efficiency and performance on various computer vision problems such as image classification
\cite{53krizhevsky2012imagenet}, object recognition
\cite{29yun2019cutmix}, video representation learning
\cite{54lin2021self}. There are also some prior works in continual learning. Ni et al.
\cite{55ni2021alleviate} combine data augmentation and contrastive learning to alleviate representation overlapping. Bang et al.
\cite{56bang2021rainbow} consider the problem of blurry task boundary where tasks share classes and then propose to apply augmentation strategy to enhance the sample diversity in the memory. In this work, we propose to mix the image-label of the current task and previous tasks randomly to augment the training data, then improve the adaptability of the model. 

\textbf{Multi-scale Learning.} Multi-scale learning has been proved great potential on various machine learning problems
\cite{57horstemeyer2009multiscale}. For example, Douillard et al.
\cite{58douillard2020plop} apply multi-scale learning to capture a higher degree of spatial precision in the field of semantic segmentation. He et al.
\cite{59he2015spatial} design a spatial pyramid pooling method to improve the performance of object detection. Recently, in continual learning, Carta et al.
\cite{60carta2020incremental} design a new incrementally trained recurrent architecture, called the MultiScale LMN to learn the short-term and long-term dependencies of the time series. Zhou et al.
\cite{61zhou2019m2kd} present a multi-model knowledge distillation strategy to further preserve knowledge encoded at the intermediate feature levels. Inspired by this, in this paper we further constraint the evolution of the long-range and short-range spatial relationships at the feature level to consolidate previous knowledge when learning new tasks in the continual learning setting. Then, we alleviate the problem due to the data imbalance.

\section{Continual Learning via the Meta-learning update with Multi-scale Knowledge Distillation and Data Augmentation (MMKDDA)}

\textbf{Problem Formulation.} The problem of continual learning is canonically formulated by the sequential training protocol
\cite{62hadsell2020embracing}. Compared with the traditional machine learning scenes of a stationary circumstance, the continual learning scenario explicitly concentrates on changing or non-stationary circumstances, often divides a set of tasks that should be learned consecutively one by one. The ultimate goal of continual learning is to maintain its performance throughout a range of all seen tasks.

Without loss of generality, we describe the problem of continual learning in an online classification scenario. For continual learning, a learner observes a sequence of labeled training samples, that arrive sequentially (one-by-one), and every sample can be expressed as a triplet $(x,\;t,\;y) \in \mathcal{X} \times N^ +   \times \mathcal{Y}$, where $x$ is an input representing the features (e.g. an image), $t$ is a task descriptor identifying a particular task(e.g. an integer), and $y$ is the class label (e.g. the specific predictive object for the input). It is noted that when the task descriptor $t$ is given, a pair $(x,y)$ can be assumed to be independent and identically distributed (IID), that is, the training sample is obtained from some probability distribution $P_t$ which is specified by the task $t$. As is shown in Fig.1, we describe a two-task learning problem.

In a real-world scenario, the agent must be able to adapt to the changing environment effectively and efficiently. Unfortunately, it is forbidden to design different models for different tasks. The goal of continual learning is to learn a function $f_\theta  \left( x \right)$, parameterized by a vector $\theta$, that means $f_\theta  \left( x \right) \approx y$, for any possible task and sample. The learner would be capable of learning the new knowledge and adapting to the changes, while also not catastrophically forgetting previously learned knowledge in a continual learning setting. This scenario requires that the learner should not forget how to perform the old tasks when learning a new task.

\begin{figure}[H]
	\centering
	\includegraphics[width=\textwidth]{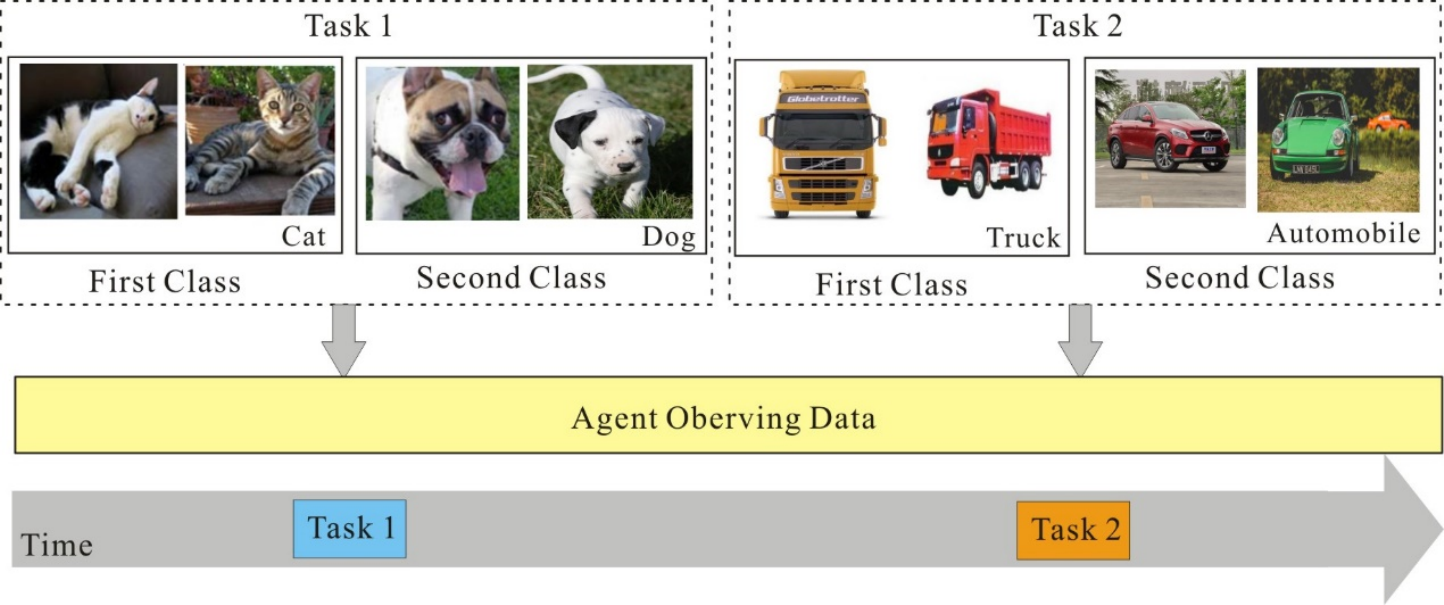}
	\caption{The paradigm for continual learning. As defined here, the tasks arrive sequentially and the agent is required to immediately learn the new task. The agent should not forget how to perform the old task after it is exposed to the new task. It is noted that the agent only has one look at each sample. It is obvious that continual learning is closer to animal learning.}
	\label{fig1}
\end{figure}

\textbf{MMKDDA Method.} In this work, we propose a framework called meta-learning via multi-scale knowledge distillation and data augmentation (MMKDDA) that can learn consecutive tasks in an online scenario. For continual learning, plasticity and stability are two pivot factors. On the one hand, higher plasticity can cause the forgetting of previous tasks while learning well on new tasks. On the other hand, higher stability can weaken the ability to perform new tasks due to the effort to maintain old knowledge. So how to effectively address the stability-plasticity dilemma is a crucial problem. 

Note that the fixed-size memory replay techniques have shown exceptional promise to combat the forgetting of old knowledge. But the limited samples from old tasks will lead to the imbalanced training distribution, which would create a bias, and exacerbate the forgetting. Meanwhile, the samples from the current task and a limited number of samples fetched from episodic memory are dictated by the different marginal distributions (i.e.$P^{current} \left( X \right) \ne P^{memory} \left( X \right)$). Obviously, in this setup, learning to adapt to the marginal distributions is very important, which would be beneficial to build a generic model with extensive applicability and practical significance. In summary, how to further optimize and balance between plasticity and stability is very important to construct a continual learning model.

Inspired by the above discussion, in this paper we propose a conceptually novel framework MMKDDA. In this framework, to rapidly adapt to the new task, the Cutmix augmentation strategy is used, which can project the samples to a more complex dimension by mixing the instance-label of the samples of the current task and the episodic memory. Furthermore, we incorporate multi-scale knowledge distillation to preserve knowledge at different feature levels to further promote consistency with past tasks. After that, an update strategy depending on the number of tasks arrived is introduced to make the right balance between plastic and stability in an online continual learning scenario. Our MMKDDA framework is shown in Fig.2. Specifically, our framework consists of three main components which are illustrated in Fig.2 a), b) and c), respectively. More precisely, (i) Construction of the training data: the current training data composes of three parts: the samples of current tasks, the samples of episodic memory from previous tasks, and the augmented data generated using the episodic memory and the current training data. (ii) Training the model: a joint training process is performed on the three parts of the training data. This training procedure involves two updates, an inner loop and an outer loop. In the inner loop, the model can quickly learn new knowledge and an outer loop would further be beneficial to form a final generic model; (iii) Building the loss function: the whole objective function consists of a cross-entropy loss $L_{{\text{BCE}}} $ and a multi-scale knowledge distillation loss $L_{{\text{multi}} - scale{\text{ KL}}}$,where $L_{{\text{multi}} - scale{\text{ KL}}} $ is a regularizer to transfer the long-range and short-range spatial relationships at the feature level.

Next, we would describe different modules in our proposed MMKDDA in detail.

\begin{figure}[H]
	\centering
	\includegraphics[width=\textwidth]{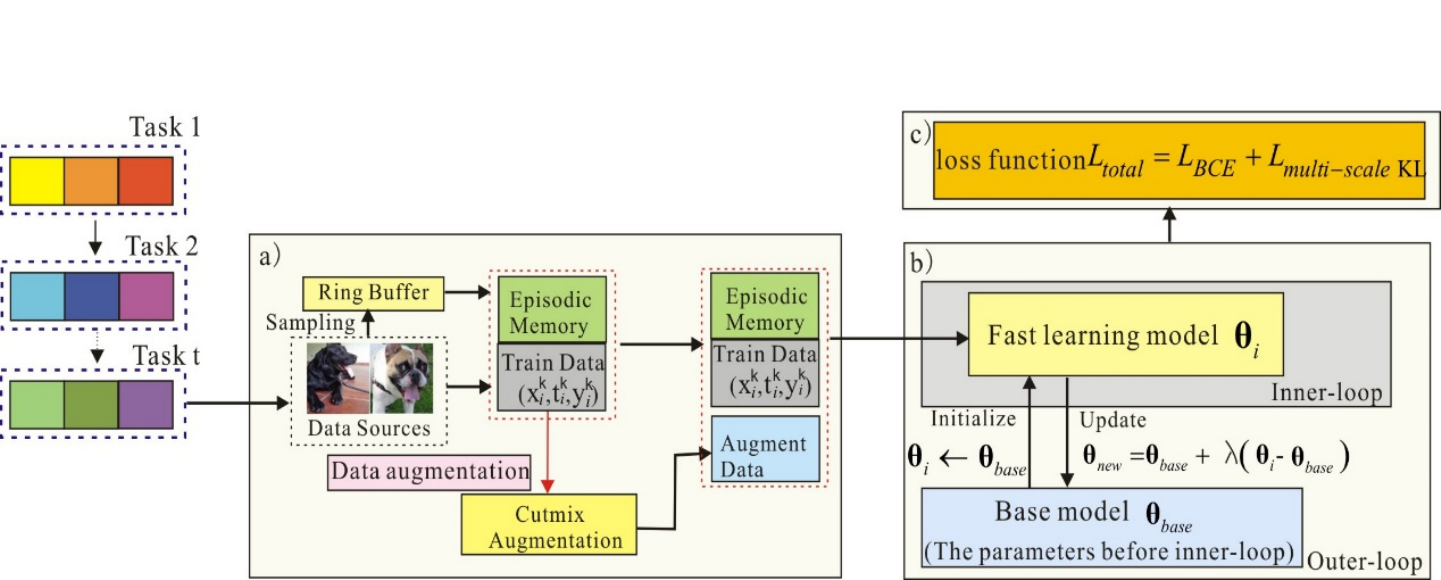}
	\caption{Illustration for our proposed MMKDDA. Firstly, on the far left side, the constitutive tasks arrive sequentially and are indexed by Task 1, Task 2, …, Task t. Panel (a) illustrates the construction of the training data. More specifically, episodic memory is used to alleviate catastrophic forgetting and we select the ring buffer strategy as our memory writing strategy. We apply the revised Cutmix strategy to generate the augmentation data. Panel (b) depicts the meta-learning update strategy based on the number of tasks seen previously. Panel (c) demonstrates the loss function.}
	\label{fig2}
\end{figure}

\subsection{Construction of the training data}

Catastrophic forgetting is an intractable problem in continual learning. Recent studies
\cite{26chaudhry2019continual,44lopez2017gradient} demonstrate that compared to regularization methods, the approaches based on episodic memory enjoy superior performance in a single epoch setting. So in this work, we utilize the experience replay (ER) strategy to alleviate catastrophic forgetting 
\cite{63chaudhry2019continual}. Moreover, we select the ring buffer strategy presented in 
\cite{26chaudhry2019continual} as our memory writing rule. However, different from the traditional ER method, an effective data augmentation strategy is emplyed to construct the training data.

\subsubsection{Memory Replay}

In this section, we introduce the experience replay algorithm in details. Algorithm 1 summarizes the experience replay strategy used in this work. Firstly, as is shown in line 8 of Algorithm 1, there is a writing strategy that is used to update the episodic memory. Then, the model parameters are updated by stacking the practical mini-batch of instances with the episodic memory by random sampling, as is shown in line 7 of Algorithm 1. Moreover, the model can avoid catastrophic forgetting by these two modification mechanisms. Algorithm 1 illustrates the full ER procedure.

\begin{table}[H]
  \centering
    \resizebox{\textwidth}{!}{
    \begin{tabular}{l}
    \toprule
    \multicolumn{1}{l}{\textbf{Algorithm 1 Experience Replay in Online Continual Learning}} \\
    \midrule
     \multicolumn{1}{l}{1: \textbf{Input}(D,$mem\_size$,$batch\_size$,$lr$)}\\
    \midrule
    \multicolumn{1}{l}{2:\quad\quad\quad$M \leftarrow \left\{ {} \right\}*mem\_size$\quad\quad\quad$\triangleright$Allocate episodic memory buffer where $mem\_size$ denotes the size of memory}\\

     \multicolumn{1}{l}{3:\quad\quad\quad$n \leftarrow 0$\quad\quad\quad$\triangleright$$n$ is the number of current training instances and initializes to 0}\\

     \multicolumn{1}{l}{4:\quad\quad\quad\quad\textbf{for $k \in \left\{ {1, \cdots ,T} \right\}$ do}}\\

     \multicolumn{1}{l}{5:\quad\quad\quad\quad\quad\textbf{for $B_n  \leftarrow D_k$ do}\quad\quad\quad$\triangleright$Obtain a mini-batch $B_n$ of training instances from the current task $k$}\\

     \multicolumn{1}{l}{6:\quad\quad\quad\quad\quad\quad$B_M  \leftarrow M$\quad\quad\quad$\triangleright$Obtain a mini-batch $B_M$ from episodic memory $M$}\\

     \multicolumn{1}{l}{7:\quad\quad\quad\quad\quad\quad$\theta  \leftarrow SGD(B_n ,B_M ,\theta ,lr)$\quad$\triangleright$Update the model parameters $\theta$ by stacking current training instances with previous instance from episodic memory $M$}\\

     \multicolumn{1}{l}{8:\quad\quad\quad\quad\quad\quad$M \leftarrow Update\_Memory\left( {mem\_size,k,n,B_n } \right)$\quad\quad\quad$\triangleright$ Update Episodic Memory}\\

     \multicolumn{1}{l}{9:\quad\quad\quad\quad\quad\quad$n \leftarrow n + batch\_size$\quad\quad\quad$\triangleright$ Update the number of training instance seen in this setting}\\

     \multicolumn{1}{l}{10:\quad\quad\quad\textbf{Return $M,\theta $}}\\
    \bottomrule
    \end{tabular}%
}
\end{table}%

\subsubsection{Data Augmentation Strategy}

In the continual learning setup, a plastic network requires that this model can quickly adapt to the desired task, which is especially important in the online scenario that the model must learn new knowledge in one epoch. We consider that different tasks enjoy the different probability distribution of the training data. Thus, when the memory replay is performed, the marginal distributions are different between the samples of the new task and a limited number of ones gained from episodic memory, i.e.,$P^{current} \left( X \right) \ne P^{memory} \left( X \right)$. An interesting question is: how can we quickly adapt to the change of probability distribution to further improve the model performance. 

When a mini-batch $B_n  \leftarrow D_k $ is observed, and in order to prevent catastrophic forgetting, we also fetch additionally a batch of a fixed number of the memory replay examples $B_M  \leftarrow M$ from episodic memory. After that, we can obtain the current mini-batch training data $B_n  \cup B_M $ composed of the current task and old tasks. Inspired by the data augmentation idea, we apply an augmentation strategy Cutmix
\cite{29yun2019cutmix} to deal with the change of probability distribution. Specifically, firstly, we shuffle the mixed mini-batch training data $B_n  \cup B_M $. Then, we remove a portion of a certain patch from an image and replace the removed regions with a patch extracted from another image. After that, we also mix the class label proportionally to the number of pixels for newly generated images. The Cutmix strategy used in this context has the following advantages. Firstly, we can obtain a new augmented mini-batch whose probability distribution is interpolated between the probability distribution of the old and current tasks. Secondly, When CutMix technology is used, the added or dropout patches would further enhance localization ability by requiring the model to identify the local properties of the object. Lastly, the localization property is also significantly enhanced by the added patches, which would make the model more sensitive to the localization characteristics of the object. 

The Cutmix approach is described in detail below. Here, let $x \in \Re ^{W \times H \times C} $ and $y$ denote an input image and the corresponding class label respectively. We can apply the Cutmix strategy to generate a novel training instance-label pair $\left( {\tilde x,\tilde y} \right)$ by combining two training instances $\left( {x_1 ,y_1 } \right)$ and $\left( {x_2 ,y_2 } \right)$. This combining operation is defined as follows,

\begin{equation}
\begin{gathered}
  \tilde x\;{\text{ = }}\;{\mathbf{M}} \odot x_1  + \left( {1 - {\mathbf{M}}} \right) \odot x_2  \hfill \\
  \tilde y\;{\text{ = }}\;\gamma y_1  + \;\left( {1 - \gamma } \right)y_2  \hfill \\ 
\end{gathered} 
\label{1}
\end{equation}

here,${\mathbf{M}} \in \left\{ {0,1} \right\}^{W \times H} $ is a binary mask, which is used to remove and fill in from two images, and the symbol $\odot $ denotes element-wise multiplication.$\gamma$ stands for the combination ratio about the two data objects which is sampled from the beta distribution ${\text{Beta}}\left( {\alpha ,\alpha } \right)$. Note that in our experiment, the parameter $\alpha$ is set as $1$, that means the combination ratio $\gamma$ is sampled from the uniform distribution $\left( {0,\;1} \right)$.

Now we describe how to generate the binary mask ${\mathbf{M}}$. Firstly, we get the bounding box coordinate ${\mathbf{B}} = \left( {r_{w1} ,\;r_{h1} ,\;r_{w2} ,\;r_{h2} } \right)$ using Eq.(2), and then we can obtain the cropping regions on $x_1 $ and $x_2 $. After that, the region ${\mathbf{B}}$ in image $x_1 $ is cropped and filled using the region ${\mathbf{B}}$ in image $x_2 $.We sample the bounding box coordinates by following rules: 

\begin{equation}
\begin{gathered}
  r_{w1} {\text{\~Uniform}}\left( {0,\;W} \right),\quad r_{w2}  = W\sqrt {1 - \gamma }  \hfill \\
  r_{h1} {\text{\~Uniform}}\left( {0,\;H} \right),\quad r_{h2}  = H\sqrt {1 - \gamma }  \hfill \\ 
\end{gathered} 
\label{2}
\end{equation}

It is obvious that the above Eq.(2) can ensure the removed region ratio $\left( {{{r_{w2} r_{h2} } \mathord{\left/
 {\vphantom {{r_{w2} r_{h2} } {WH}}} \right.
 \kern-\nulldelimiterspace} {WH}}} \right){\text{ = }}1 - \gamma $. During the training process, we generate the new mix-ed instances by mixing the two training instances in mini-batch training data $B_n  \cup B_M $ using Eq.(1). It is indicated that the Cutmix strategy replaces an image region with a patch from another training image and generates a set of images whose probability distribution is between the probability distribution of the current task’s instances and the previous task’s instances. It is helpful to alleviate the influence caused by the change of probability distribution over the tasks and enforce the model to learn new knowledge from current instances quickly while not forgetting the old knowledge. 

The complete algorithm of Cutmix is provided in Algorithm 2.

\begin{table}[H]
  \centering
    \resizebox{\textwidth}{!}{
    \begin{tabular}{l}
    \toprule
    \multicolumn{1}{l}{\textbf{Algorithm 2} Cutmix strategy is used in online continual learning} \\
    \midrule
     \multicolumn{1}{l}{1: \textbf{Required:}Current minibatch training data $\left( {{\mathbf{x}}_i ,\;{\mathbf{y}}_i } \right)\triangleright{\mathbf{x}}_i$is tensor with $N \times C \times W \times H$ size,${\mathbf{y}}_i $ is tensor with $N \times K$ size}\\

    \multicolumn{1}{l}{2:\quad\quad\quad Shuffle the training data $\left( {{\mathbf{x}}_i^{shuffle} ,\;{\mathbf{y}}_i^{shuffle} } \right){\text{ = Shuffle\_minibatch}}\left( {{\mathbf{x}}_i ,\;{\mathbf{y}}_i } \right)$}\\

     \multicolumn{1}{l}{3:\quad\quad\quad Sample the parameter $\gamma$: $\gamma {\text{ = Uniform}}\left( {0,1} \right)$}\\

     \multicolumn{1}{l}{4:\quad\quad\quad Sample $r_{w1}$, $r_{h1}$: $r_{w1} {\text{ = Uniform}}\left( {0,\;W} \right);{\text{  }}r_{h1} {\text{ = Uniform}}\left( {0,\;H} \right)$}\\

     \multicolumn{1}{l}{5:\quad\quad\quad Compute $r_{w2}$, $r_{h2}$: $r_{w2}  = W\sqrt {1 - \gamma } {\text{,  }}r_{h2}  = H\sqrt {1 - \gamma }$}\\

     \multicolumn{1}{l}{6:\quad\quad\quad Compute the cropping region:$w1 = {\text{Round}}\left( {{\text{Clip}}\left( {r_{w1}  - {\raise0.7ex\hbox{${r_{w2} }$} \!\mathord{\left/
 {\vphantom {{r_{w2} } 2}}\right.\kern-\nulldelimiterspace}
\!\lower0.7ex\hbox{$2$}},\;\min  = 0} \right)} \right);\;$}\\

     \multicolumn{1}{l}{\quad\quad\quad\quad\quad\quad
\quad\quad\quad\quad\quad\quad
\quad\quad\quad\quad\quad $w2 = {\text{Round}}\left( {{\text{Clip}}\left( {r_{w1}  + {\raise0.7ex\hbox{${r_{w2} }$} \!\mathord{\left/
 {\vphantom {{r_{w2} } 2}}\right.\kern-\nulldelimiterspace}
\!\lower0.7ex\hbox{$2$}},\;\min  = W} \right)} \right)$}\\

    \multicolumn{1}{l}{\quad\quad\quad\quad\quad\quad
\quad\quad\quad\quad\quad\quad
\quad\quad\quad\quad\quad $h1 = {\text{Round}}\left( {{\text{Clip}}\left( {r_{h1}  - {\raise0.7ex\hbox{${r_{h2} }$} \!\mathord{\left/
 {\vphantom {{r_{h2} } 2}}\right.\kern-\nulldelimiterspace}
\!\lower0.7ex\hbox{$2$}},\;\min  = 0} \right)} \right)$}\\

    \multicolumn{1}{l}{\quad\quad\quad\quad\quad\quad
\quad\quad\quad\quad\quad\quad
\quad\quad\quad\quad\quad $h2 = {\text{Round}}\left( {{\text{Clip}}\left( {r_{h1}  + {\raise0.7ex\hbox{${r_{h2} }$} \!\mathord{\left/
 {\vphantom {{r_{h2} } 2}}\right.\kern-\nulldelimiterspace}
\!\lower0.7ex\hbox{$2$}},\;\min  = H} \right)} \right)$}\\

     \multicolumn{1}{l}{7:\quad\quad\quad Mix the ${\mathbf{x}}_i$ and ${\mathbf{x}}_i^{shuffle}$ to generate ${\mathbf{\tilde x}}$: ${\mathbf{x}}_i \left[ {:,\;:,\;w1:w2,\;h1:h2} \right] = {\mathbf{x}}_i^{shuffle} \left[ {:,\;:,\;w1:w2,\;h1:h2} \right]$}\\

     \multicolumn{1}{l}{8:\quad\quad\quad Compute the exact region ratio: $\gamma  = 1 - \left( {w2 - w1} \right) \cdot {{\left( {h2 - h1} \right)} \mathord{\left/
 {\vphantom {{\left( {h2 - h1} \right)} {\left( {W \times H} \right)}}} \right.
 \kern-\nulldelimiterspace} {\left( {W \times H} \right)}}$}\\

     \multicolumn{1}{l}{9:\quad\quad\quad Gain the target label: ${\mathbf{\tilde y}}\;{\text{ = }}\;\gamma  \cdot {\mathbf{y}}_i  + \left( {1 - \gamma } \right) \cdot {\mathbf{y}}_i^{shuffle} $}\\

     \multicolumn{1}{l}{10:\quad\quad\quad Generate new augmentation data $\left( {{\mathbf{\tilde x}},{\mathbf{\tilde y}}\;} \right)$}\\

     \multicolumn{1}{l}{11: \textbf{end}}\\
    \bottomrule
    \end{tabular}%
}
\end{table}%

\subsection{Meta-learning update based on the number of tasks seen previously}

It is important to keep the right balance between stability and plasticity. Thus, in this paper, we introduce a new meta update strategy based on the number of tasks seen previously where the meta-learning rate is updated via the dynamic controller method
\cite{64rajasegaran2020itaml}. The monotonic scheduling strategy of the trade-off parameter of old and new tasks would enforce the model to learn the inherent relationships between consecutive tasks. Thus, this update strategy would be beneficial to obtain a more generic meaningful representation to fight against the stability-plasticity dilemma. So, in the following part, we elaborate on the trade-off rule in detail.

We progressively learn $T$ sequential tasks, where ${\mathbf{\theta }}$ denotes the model parameters. Our meta-learning update approach involves two-loop updates: an inner-loop update and an outer-loop update. Here, an inner-loop update can learn a task-specific model for each task, which can quickly adapt to new knowledge. An outer-loop update would further make a balance between stability and plasticity and then form a final generic model. 

\textbf{Inner-loop update:}	To perform the inner-loop update, firstly we observe a mini-batch composed of the current task training data and the samples gotten from the episodic memory. Then, we train the model by minimizing the loss function (e.g. binary cross-entropy loss) to obtain our inner-loop parameters ${\mathbf{\theta }}_i$, and the loss for temporary model is computed via SGD optimizer according to:

\begin{equation}
{\mathbf{\theta }}_i^t {\text{ = SGD}}^k {\text{(}}{\mathbf{\theta }}_{{\text{base}}} ,B_n ,\beta {\text{)}}
\label{3}
\end{equation}

where $\beta $ denotes the inner learning rate,${\mathbf{\theta }}_{{\text{base}}}$ denotes the model parameters generated before inner-loop updates and $k$ denotes the number of updates in the set of current training instances $B_n$($k$ is set to 2 in our all experiments). After every episode, the final model parameters can be obtained according to the following outer-loop update.

\textbf{Outer-loop update:}	In the outer-loop, we combine the inner-loop parameters ${\mathbf{\theta }}_i$ to form the final model parameters $\theta$. Here, we apply ${\mathbf{\theta }}_{{\text{base}}}$ to denote the model parameters generated before inner-loop updates. Then $\left( {{\mathbf{\theta }}_i  - {\mathbf{\theta }}_{{\text{base}}} } \right)$ denotes the gradient change after the outer-loop is performed
\cite{65nichol2018first}

After that, we apply a monotonic scheduling strategy via a dynamic controller $\lambda$ to control the evolution of the gradient, which can dynamically enforce the changes of the parameters towards the average direction of all tasks update. Finally, we can achieve the trade-off between the stability-plasticity. The update rule is given as follow:

\begin{equation}
{\mathbf{\theta }}_{{\text{new}}} \;{\text{ = }}\;{\mathbf{\theta }}_{{\text{base}}}  + \lambda \left( {{\mathbf{\theta }}_i  - {\mathbf{\theta }}_{{\text{base}}} } \right){\text{  where  }}\lambda  = e^{\left( { - \rho  \cdot \frac{t}
{T}} \right)} 
\label{4}
\end{equation}

The monotonic scheduling strategy of the trade-off parameter of old and new tasks can dictate the model to learn the inherent relationship between sequential tasks. Specifically, the model speeds up the learning at the beginning and slows down as the training progress, because ${\mathbf{\theta }}_{{\text{base}}}$ contains knowledge for more old tasks as the number of tasks observed increases. The balancing factor is defined as $\lambda  = e^{\left( { - \rho  \cdot \frac{t}
{T}} \right)} $, where $t$ and $T$ denote the current task and the total number of tasks respectively and $\rho$ denotes the decay rate. For the outer-loop trade-off rule, the decay rate $\rho$ is set to 2, 3, 2 and 1 for Split CIFAR10, Split SVHN, Split CIFAR100, and Split TinyImagenet200 respectively. The controller depends on the number of tasks observed. Thus, the model can learn the inherent relationship between sequential tasks by this the monotonic scheduling strategy, which is beneficial to make a better trade-off between plasticity and stability.

The complete algorithm about meta-learning update based on the number of tasks seen previously is provided in Algorithm 3.

\begin{table}[H]
  \centering
   \resizebox{\textwidth}{!}{
    \begin{tabular}{l}
    \toprule
    \multicolumn{1}{l}{\textbf{Algorithm 3} The Meta-learning update based on the number of tasks seen previously\quad\quad\quad\quad\quad\quad\quad\quad\quad\quad\quad\quad\quad\quad\quad} \\
    \midrule

    \multicolumn{1}{l}{1: \textbf{Required:} ${\mathbf{\theta }}_{{\text{base}}}$, ${\text{input}}\;{\text{ = }}\;B_n  \cup B_M $, $t$, $T$}\\

    \multicolumn{1}{l}{2: \textbf{for} each iteration \textbf{do}}\\

    \multicolumn{1}{l}{3:\quad\quad\quad ${\mathbf{\theta }}_i  \leftarrow {\mathbf{\theta }}_{{\text{base}}}$}\\

    \multicolumn{1}{l}{4:\quad\quad\quad Get the output of model:\quad\quad${\text{output}}\;{\text{ = }}\;{\mathbf{\theta }}_{{\text{base}}} \left( {{\text{input}}} \right)$}\\

    \multicolumn{1}{l}{5:\quad\quad\quad Compute loss:\quad\quad${\text{loss}}\;{\text{ = }}\;{\text{loss\_function}}\left( {{\text{output,}}\;{\text{target}}} \right)$}\\

    \multicolumn{1}{l}{6:\quad\quad\quad Obtain the inner-loop parameters:\quad\quad${\mathbf{\theta }}_i  = {\text{Optimizer}}\left( {{\mathbf{\theta }}_i ,\;{\text{loss}}} \right)$}\\

    \multicolumn{1}{l}{7:\quad\quad\quad Compute the balancing factor:\quad\quad$\lambda  = e^{\left( { - \rho  \cdot \frac{t}
{T}} \right)}$}\\

    \multicolumn{1}{l}{8:\quad\quad\quad Compute the model parameters:\quad\quad${\mathbf{\theta }}\;{\text{ = }}\;{\mathbf{\theta }}_{{\text{base}}}  + \lambda \left( {{\mathbf{\theta }}_i  - {\mathbf{\theta }}_{{\text{base}}} } \right)$}\\

    \multicolumn{1}{l}{9:\quad\quad\quad \textbf{return ${\mathbf{\theta }}$}}\\

    \multicolumn{1}{l}{10: \textbf{end}}\\

    \bottomrule
    \end{tabular}%
}
\end{table}%

\subsection{Building the loss function}

In continual learning scenarios, agents must fight against the catastrophic forgetting of old knowledge to preserve the model stability. However, when the memory replay technique is performed, the imbalance between new and old tasks’ training data would aggravate the forgetting of old knowledge which is harmful to maintain the model stability. Thus, in order to alleviate this issue and inspired from the multi-scale literature
\cite{58douillard2020plop,59he2015spatial}, \cite{66lazebnik2006beyond}, we apply a multi-scale knowledge distillation approach to constrain the evolution of long-range and short-range spatial relationships at the feature level, which would be more beneficial to model details of the image, especially in the online setting. Besides, this method can yield fruitful results because of its ability to regulate the evolution of representation at different-level layers. So, we regularize the training loss by applying multi-scale knowledge distillation. Let $\hat y_k $ denote the logits of the model’s prediction and $y_k $ is the corresponding truth label.

\begin{equation}
L_{total} \;{\text{ = }}\;L_{BCE} \left( {\hat y_k ,y_k \;} \right) + \gamma L_{{\text{multi}} - scale{\text{ KL}}} \left( {{\mathbf{h}}{\kern 1pt} ^t ,{\mathbf{h}}{\kern 1pt} ^{t{\text{ - }}1} } \right),
\label{5}
\end{equation}

where $L_{BCE} $ is the cross-entropy loss,$\gamma $ is the trade-off parameter ($\gamma $ is set to 0.1), and $L_{{\text{multi - }}scale{\text{ KL}}}$ is the multi-scale knowledge distillation regularizer. It is worth noting that the superscript $t$ denotes the model learned at current task $t$. ${\mathbf{h}}^t$ and ${\mathbf{h}}^{t{\text{ - }}1} $ denote the intermediate feature layer at task $t$ and $t{\text{ - }}1$ respectively. In the following section, we would present the multi-scale knowledge distillation in detail.

\textbf{Multi-scale Knowledge Distillation.}  Specifically, this distillation process mainly consists of two steps: (i) firstly, we compute width and height slices on multiple regions extracted at various scales $\left\{ {{{1{\kern 1pt} } \mathord{\left/
 {\vphantom {{1{\kern 1pt} } {{\kern 1pt} 2^{{\kern 1pt} s} }}} \right.
 \kern-\nulldelimiterspace} {{\kern 1pt} 2^{{\kern 1pt} s} }}} \right\}_{s{\kern 1pt}  = {\kern 1pt} 0{\kern 1pt} ,\; \cdots \;,\;S} {\kern 1pt} $. (ii) then we match the output of global statistics at different feature levels between the current and previous models. This process is illustrated in Fig.3. 

\begin{figure}[H]
	\centering
	\includegraphics[width=\textwidth]{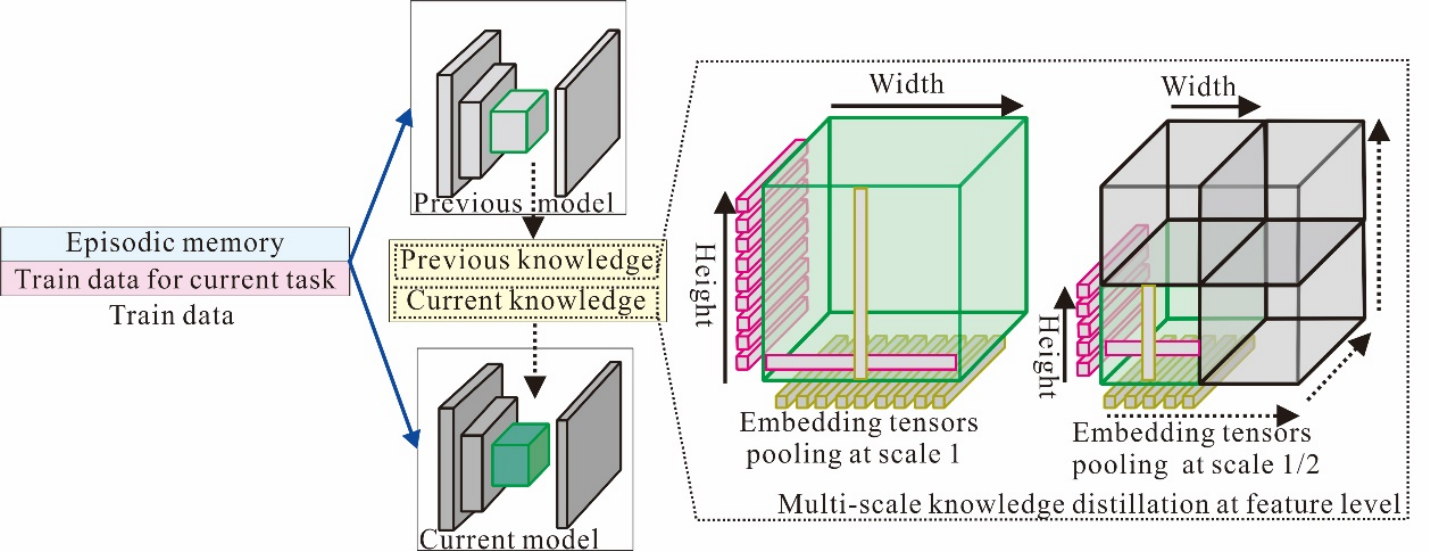}
	\caption{Illustration of multi-scale knowledge distillation. Here, we employ the multi-scale knowledge distillation and obtain the different embedding tensors at scales 1 and ${{\text{1}} \mathord{\left/
 {\vphantom {{\text{1}} {\text{2}}}} \right.
 \kern-\nulldelimiterspace} {\text{2}}}$ , respectively.}
	\label{fig3}
\end{figure}

\textbf{Feature extraction with multiple regions.} Without loss of generality, for the sake of simplicity, we mainly introduce a relatively simple case where the embedding tensors are considered at scales 1 and ${{\text{1}} \mathord{\left/
 {\vphantom {{\text{1}} {\text{2}}}} \right.
 \kern-\nulldelimiterspace} {\text{2}}}$ respectively. Here, firstly let us consider a simple case at scale $s\;{\text{ = }}\;1$. Assume that we progressively learn a sequence of $T$ tasks. Normally, a classification model at step $t$ can be regarded as two sub-nets, a subnetwork for extracting feature $f^t \left(  \cdot  \right)$ and a classification subnetwork $g^t \left( {f\;^t \left(  \cdot  \right)} \right)$. By this architecture, we can extract the different-level features ${\mathbf{h}}{\kern 1pt} _l^t \; = \;f_l^t \left(  \cdot  \right)\,,\,\,l \in \left\{ {1{\kern 1pt} \,,\, \cdots \,,\,L} \right\}$ at different layer $l$ . Let $\hat y\,\; = \;g^t \left( {{\mathbf{h}}_L^t } \right)$ denote the final prediction output. ${\mathbf{h}}{\kern 1pt} ^t $ is the embedding tensor of size $H \times W \times C$ for the intermediate feature layer. Here, an embedding feature can be obtained by concatenating the $W \times C$ height-pooled slices and the $H \times C$ width-pooled slices for ${\mathbf{h}}{\kern 1pt} _l^t \;$. The formal expression is given by:

\begin{equation}
\Phi \left( {{\mathbf{h}}{\kern 1pt} _l^t } \right){\text{ = }}\left[ {\frac{1}
{W}\sum\limits_{w = 1}^W {{\mathbf{h}}{\kern 1pt} _l^t \left[ {:,w,:} \right]\left\| {\frac{1}
{W}\sum\limits_{w = 1}^W {{\mathbf{h}}{\kern 1pt} _l^t \left[ {h,:,:} \right]} } \right.} } \right]\; \in \Re ^{\left( {H + W} \right) \times C} 
\label{6}
\end{equation}

here, $\left[ { \cdot \left\|  \cdot  \right.} \right]$ denotes concatenation over the channel axis. It is noted that for our approach, these embedding tensors are computed by the above expression at several layers for both the current and previous task model. After that, the knowledge distillation loss at various feature levels is computed by minimizing the Euclidean distance between the two sets of embedding over the new and old model:

\begin{equation}
L_{kl}  = \frac{1}
{L}\sum\limits_{l = 1}^L {\left\| {\Phi \left( {{\mathbf{h}}{\kern 1pt} _l^t } \right){\text{ - }}\Phi \left( {{\mathbf{h}}{\kern 1pt} _l^{t{\text{ - }}1} } \right)} \right\|^2 } 
\label{7}
\end{equation}

Similar to the above at scale $s\;{\text{ = }}\;1$, the embedding tensor $\Psi ^s \left( {{\mathbf{h}}{\kern 1pt} ^t } \right)$ at scale $s$ is calculated as the concatenation of $s^2$ embedding features (at scale $s\;{\text{ = }}\;0$). This specific formula is given by:

\begin{equation}
\Psi ^s \left( {{\mathbf{h}}{\kern 1pt} ^t } \right){\text{ = }}\left[ {\Phi \left( {{\mathbf{h}}{\kern 1pt} _{0,0}^{t,\,s} } \right)\left\| {\; \cdots \;} \right\|\Phi \left( {{\mathbf{h}}{\kern 1pt} _{s - 1,s - 1}^{t,\,s} } \right)} \right] \in \Re ^{\left( {H + W} \right) \times C} 
\label{8}
\end{equation}

here, for $\forall i = 0 \cdots s - 1$, and $\forall j = 0 \cdots s - 1$,

\begin{equation}
\;{\mathbf{h}}{\kern 1pt} _{i,j}^{t,\,s}  = {\mathbf{h}}{\kern 1pt} ^{t,s} \left[ {{{iH} \mathord{\left/
 {\vphantom {{iH} s}} \right.
 \kern-\nulldelimiterspace} s}:{{\left( {i + 1} \right)H} \mathord{\left/
 {\vphantom {{\left( {i + 1} \right)H} {s,{{\;jW} \mathord{\left/
 {\vphantom {{\;jW} s}} \right.
 \kern-\nulldelimiterspace} s}:{{\left( {j + 1} \right)W} \mathord{\left/
 {\vphantom {{\left( {j + 1} \right)W} s}} \right.
 \kern-\nulldelimiterspace} s}}}} \right.
 \kern-\nulldelimiterspace} {s,{{\;jW} \mathord{\left/
 {\vphantom {{\;jW} s}} \right.
 \kern-\nulldelimiterspace} s}:{{\left( {j + 1} \right)W} \mathord{\left/
 {\vphantom {{\left( {j + 1} \right)W} s}} \right.
 \kern-\nulldelimiterspace} s}}}} \right]
\label{9}
\end{equation}

denotes a sub-region with size ${{\;W} \mathord{\left/
 {\vphantom {{\;W} s}} \right.
 \kern-\nulldelimiterspace} s} \times {H \mathord{\left/
 {\vphantom {H s}} \right.
 \kern-\nulldelimiterspace} s}$ at scale $s$. After that, the final embedding tensor is obtained by concatenating the embeddings for multiple regions $\Psi ^s \left( {{\mathbf{h}}{\kern 1pt} ^t } \right)$ with different scale $s$, where $s \in 0,\; \cdots ,\;S$, i.e.,

\begin{equation}
\Psi \left( {{\mathbf{h}}{\kern 1pt} ^t } \right){\text{ = }}\left[ {\Psi ^1 \left( {{\mathbf{h}}{\kern 1pt} ^t } \right)\left\| {\; \cdots \;} \right\|\Psi ^S \left( {{\mathbf{h}}{\kern 1pt} ^t } \right)} \right] \in \Re ^{\left( {H + W} \right) \times C \times S} 
\label{10}
\end{equation}

we then compute the final embedding tensor of both current and previous models by minimizing $L_2$ norm distance to form the multi-scale distillation loss, and this is shown in the following formula: 

\begin{equation}
L_{{\text{multi}} - scale{\text{ KL}}} {\text{ = }}\frac{1}
{L}\sum\limits_{l = 1}^L {\left\| {\Psi \left( {{\mathbf{h}}{\kern 1pt} ^t } \right){\text{ - }}\Psi \left( {{\mathbf{h}}{\kern 1pt} ^{t{\text{ - }}1} } \right)} \right\|^2 } 
\label{11}
\end{equation}

here, the first scale $s = \left( {{{1{\kern 1pt} } \mathord{\left/
 {\vphantom {{1{\kern 1pt} } {{\kern 1pt} 2^{{\kern 1pt} 0} }}} \right.
 \kern-\nulldelimiterspace} {{\kern 1pt} 2^{{\kern 1pt} 0} }}} \right)$ can model the relationship of long-range dependencies in the context of continual learning. The subsequent scales $s = \left\{ {{{1{\kern 1pt} } \mathord{\left/
 {\vphantom {{1{\kern 1pt} } {{\kern 1pt} 2^{{\kern 1pt} 1} ,}}} \right.
 \kern-\nulldelimiterspace} {{\kern 1pt} 2^{{\kern 1pt} 1} ,}}{{1{\kern 1pt} } \mathord{\left/
 {\vphantom {{1{\kern 1pt} } {{\kern 1pt} 2^{{\kern 1pt} 2} ,}}} \right.
 \kern-\nulldelimiterspace} {{\kern 1pt} 2^{{\kern 1pt} 2} ,}} \cdots } \right\}$ can abstract short-range dependencies. In this work, we only select the subsequent scale as $\left( {{{1{\kern 1pt} } \mathord{\left/
 {\vphantom {{1{\kern 1pt} } {{\kern 1pt} 2^{{\kern 1pt} 1} }}} \right.
 \kern-\nulldelimiterspace} {{\kern 1pt} 2^{{\kern 1pt} 1} }}} \right)$ . Thus, the multi-scale knowledge distillation can effectively alleviate catastrophic forgetting due to the imbalance of training data by retaining the information of both long-range and short-range spatial relationships. 

The whole algorithm of embedding extraction for multiple regions is depicted in Algorithm 4. This algorithm composes of three main functions: multi-scale knowledge distillation, $L_2 $ norm distance between embeddings for multiple regions of the current model and the previous model, and the calculations of embedded tensors at different scale. More specifically, firstly, we employ FUNCTION 1 to obtain multi-scale knowledge distillation loss $L_{{\text{multi}} - scale{\text{ KL}}} $ by looping over all $L$ intermediate convolutional layers. Secondly, we apply $L_2 $ norm distance (Line 21) to acquire the embeddings for multiple regions of the current and previous model (Line 17). Thirdly, we calculate the embedding tensors for multiple regions given two intermediate feature maps for sub-regions (Line 16), and the embedding tensors for multiple regions at each scale $s,\;s \in \left( {0, \cdots ,S} \right)$ are obtained by looping over $S$ various scales (Line 12).

\begin{table}[H]
  \centering
   \resizebox{\textwidth}{!}{
    \begin{tabular}{l}
    \toprule
    \multicolumn{1}{l}{\textbf{Algorithm 4} Feature extraction for multiple regions
\quad\quad\quad\quad\quad\quad\quad\quad\quad\quad\quad
\quad\quad\quad\quad\quad\quad\quad\quad\quad\quad\quad
\quad\quad\quad\quad\quad\quad\quad\quad\quad\quad} \\
    \midrule

    \multicolumn{1}{l}{1: Multi-scale knowledge distillation \textbf{FUNCTION 1 $\left( {{\mathbf{h}}{\kern 1pt} ^t ,{\mathbf{h}}{\kern 1pt} ^{t{\text{ - }}1} ,S} \right)$}}\\

    \multicolumn{1}{l}{2:\quad\quad\quad loss $L_{{\text{multi}} - scale{\text{ KL}}}  \leftarrow 0$}\\

    \multicolumn{1}{l}{3: \textbf{for} $l \leftarrow 0;\;l < L;l +  + $ \textbf{do}}\\

    \multicolumn{1}{l}{4:\quad\quad\quad\quad $L_{{\text{multi}} - scale{\text{ KL}}}  + {\text{ = }}L_{{\text{multi}} - scale{\text{ KL}}} \left( {{\mathbf{h}}{\kern 1pt} ^t ,{\mathbf{h}}{\kern 1pt} ^{t{\text{ - }}1} ,S} \right)$}\\

    \multicolumn{1}{l}{5:\quad\quad\quad end for}\\

    \multicolumn{1}{l}{6:\quad\quad\quad \textbf{return} $L_{{\text{multi}} - scale{\text{ KL}}} $}\\

    \multicolumn{1}{l}{7: \textbf{end}}\\

    \multicolumn{1}{l}{8:}\\

    \multicolumn{1}{l}{}\\

    \multicolumn{1}{l}{9: \textbf{FUNCTION 2} $L_{{\text{multi}} - scale{\text{ KL}}} \left( {{\mathbf{h}}{\kern 1pt} ^t ,{\mathbf{h}}{\kern 1pt} ^{t{\text{ - }}1} ,S} \right)$} \\

    \multicolumn{1}{l}{}\\

    \multicolumn{1}{l}{10:\quad\quad\quad embedding tensors of multiple regions for current model $Le^t  \leftarrow [\;]$}\\

    \multicolumn{1}{l}{11:\quad\quad\quad embedding tensors of multiple regions for previous model $Le^{t{\text{ - }}1}  \leftarrow [\;]$}\\

    \multicolumn{1}{l}{12:\quad\quad\quad \textbf{for} $i \leftarrow 0;\;i < W - w;i +  = w$ \textbf{do}}\\

    \multicolumn{1}{l}{13:\quad\quad\quad\quad $w \leftarrow {\raise0.7ex\hbox{$W$} \!\mathord{\left/
 {\vphantom {W {2^s }}}\right.\kern-\nulldelimiterspace}
\!\lower0.7ex\hbox{${2^s }$}};\;\;h \leftarrow {\raise0.7ex\hbox{$H$} \!\mathord{\left/
 {\vphantom {H {2^s }}}\right.\kern-\nulldelimiterspace}
\!\lower0.7ex\hbox{${2^s }$}}$}\\

    \multicolumn{1}{l}{14:\quad\quad\quad\quad \textbf{for} $i \leftarrow 0;\;i < W - w;i +  = w$ \textbf{do}}\\

    \multicolumn{1}{l}{15:\quad\quad\quad\quad\quad \textbf{for} $j \leftarrow 0;\;j < H - h;j +  = h$ \textbf{do}}\\

    \multicolumn{1}{l}{16:\quad\quad\quad\quad\quad $Le^t  \leftarrow \Phi \left( {{\mathbf{h}}{\kern 1pt} ^t \,\left[ {i:i + h,j:j + h} \right]} \right)$; $Le^{t{\text{ - }}1}  \leftarrow \Phi \left( {{\mathbf{h}}{\kern 1pt} ^{t{\text{ - }}1} \,\left[ {i:i + h,j:j + h} \right]} \right)$\quad\quad\quad\quad\quad$\triangleright$Eq.5}\\

    \multicolumn{1}{l}{17:\quad\quad\quad\quad\quad $Le^{t{\text{ - }}1} \left\| {\,{\text{ = }}} \right.Le^{t{\text{ - }}1} $}\\

    \multicolumn{1}{l}{18:\quad\quad\quad\quad\quad \textbf{end for}}\\

    \multicolumn{1}{l}{19:\quad\quad\quad\quad \textbf{end for}}\\

    \multicolumn{1}{l}{20:\quad\quad\quad\ \textbf{end for}}\\

    \multicolumn{1}{l}{21:\quad\quad\quad\ \textbf{return} $\left\| {Le^t {\text{ - }}Le^{t{\text{ - }}1} } \right\|^2 $\quad\quad\quad\quad\quad\quad\quad\quad\quad\quad$\triangleright$Eq.9}\\

    \multicolumn{1}{l}{22: \textbf{end}}\\

    \bottomrule
    \end{tabular}%
}
\end{table}%

\subsection{The Specific Implementation of Our Proposed MMKDDA Framework}

In this subsection, we will describe our proposed framework. MMKDDA approach aims to alleviate catastrophic forgetting while improving the learning ability of the model. Specifically, we select the cross-entropy loss $L_{BCE} \left( {x_k ,y_k ,M_k } \right)$ as our loss function where $M_k$ denotes episodic memory which is used to overcome forgetting. When experience replay is performed, compared with the current training data, the training data from the previous task is limit which would aggravate forgetting. To overcome this problem, we apply a multi-scale knowledge distillation approach to constrain the long-range and short-range spatial relationships at different feature levels and the distillation loss $L_{{\text{multi}} - scale{\text{ KL}}} $ is obtained by measuring the difference between the current and previous task’s model. Furthermore, in order to quickly adapt to the desired task, an effective data augmentation strategy is performed. Besides, meta-learning is applied to make the model learn ‘how to rapidly learn’ and an update strategy base on the inherent relationships between sequential tasks is employed to make a great balance between stability and plasticity. Finally, the whole objective function is written as follows: 

\begin{equation}
L_{total} \;{\text{ = }}\;L_{BCE}  + \gamma L_{{\text{multi}} - scale{\text{ KL}}}  
\label{12}
\end{equation}

Obviously, our loss consists of a combination of cross-entropy loss and a multi-scale knowledge distillation loss.$\gamma $ is a tradeoff coefficient to control the influence of the old knowledge obtained in previous tasks. Our proposed MMKDDA is illustrated in the following Algorithm 5.

\begin{table}[H]
  \centering
    \resizebox{\textwidth}{!}{
    \begin{tabular}{l}
    \toprule
    \multicolumn{1}{l}{\textbf{Algorithm 5} MMKDDA algorithm} \\
    \midrule

     \multicolumn{1}{l}{1: \textbf{Required:} ${\mathbf{\theta }}_{{\text{base}}} $, ${\text{input}}\;{\text{ = }}\;B_n  \cup B_M $, $t$, $T$, $M$}\\

     \multicolumn{1}{l}{2: \textbf{for} each iteration \textbf{do}}\\

     \multicolumn{1}{l}{3: \quad\quad\quad if do Cutmix Augmentation:}\\

     \multicolumn{1}{l}{4: \quad\quad\quad\quad\quad Generate augmentation data $B_{aug}$ by Algorithm 2.  Generate mixed data: ${\text{input}}\;{\text{ = }}\;B_n  \cup B_M  \cup B_{aug} $}\\

     \multicolumn{1}{l}{5: \quad\quad\quad Get the output of model: ${\text{output}}\;{\text{ = }}\;{\mathbf{\theta }}_{{\text{base}}} \left( {{\text{input}}} \right)$}\\

     \multicolumn{1}{l}{6: \quad\quad\quad Compute loss by the Eq.(10)}\\

     \multicolumn{1}{l}{7: \quad\quad\quad Obtain the inner-loop parameters: ${\mathbf{\theta }}_i  = {\text{Optimizer}}\left( {{\mathbf{\theta }}_i ,\;{\text{loss}}} \right)$}\\

     \multicolumn{1}{l}{8: \quad\quad\quad  Obtain the final parameters ${\mathbf{\theta }}$ by Algorithm 3}\\

     \multicolumn{1}{l}{9: \quad\quad\quad Update episodic memory by Ring buffer strategy}\\

     \multicolumn{1}{l}{10: \quad\quad\quad \textbf{return} ${\mathbf{\theta }}$,and $M$}\\

     \multicolumn{1}{l}{11: \textbf{end}}\\

    \bottomrule
    \end{tabular}%
}
\end{table}%

\section{Experiments}

In this section, we first present the characteristic of the four real datasets, and then we introduce the evaluation metrics, baselines, and implementation details. We follow the standard evaluation metrics from the relative works and report results via 5-fold cross-validation. We conduct comprehensive experiments to validate the effectiveness of our proposed MMKDDA, and extensive ablation studies illustrate the impact of different factors involved in continual learning and compare with the other state-of-the-art methods. 

\subsection{Datasets}

We use the following benchmark datasets to evaluate our proposed MMKDDA in this paper.

\begin{itemize}

\item[1)]	\textbf{Split CIFAR10.} We split the CIFAR10 dataset
\cite{67krizhevsky2009learning} into 5 disjoint tasks and each task contains two classes. All images are processed into $32 \times 32$ pixels for CIFAR and there is no intersection between different tasks.

\item[2)]	\textbf{Split SVHN
\cite{64rajasegaran2020itaml}.} This dataset contains a total of 600,000 digit images with the size of $32 \times 32$ pixels from Google Street View
\cite{68netzer2011reading}, which is remarkably harder to learn. We also split this dataset into 5 disjoint tasks and each task contains two classes.

\item[3)]	\textbf{Split CIFAR100
\cite{44lopez2017gradient}.} CIFAR100 dataset
\cite{67krizhevsky2009learning} is a common object dataset that contains 100 categories and each category is composed of 600 images. In this work, we divide this dataset into 20 disjoint subsets respectively and each subset is regarded as a separated task, and then, we evaluate our MMKDDA on split CIFAR100 dataset.

\item[4)]	\textbf{4)	Split TinyImagenet200
\cite{69gupta2020maml}.} Split TinyImagenet200 is a variant of the Tiny-ImageNet collected from Flickr and other search engines. It contains 200 classes in total and we split 200 classes into 40 5-way classification tasks.

\end{itemize}

\subsection{Evaluation Metrics}

In this subsection, we introduce five evaluation metrics: Average Accuracy(ACC)
\cite{44lopez2017gradient}, Forgetting Measure(FM)
\cite{33chaudhry2018riemannian}, Backward Transfer(BWT)
\cite{44lopez2017gradient}, Forward Transfer(FWT)
\cite{44lopez2017gradient}, Learning Accuracy(LA)
\cite{27riemer2018learning}, which are used to evaluate the performance of our proposed MMKDDA and other baselines. Online continual learning considers a sequential environment where the task is observed one by one. We assume that when the task $t_i$ is finished, we measure the model performance on the test data about all T tasks. We build the matrix $a_{i,j}$, where $a_{i,j}$ represents the accuracy on the test data of the task $t_j $ after learning the last example about the current task $t_i $.$\bar b_i {\text{ }}$ is test accuracy for the task $t_i $ by random initialization. Then, the ACC, FM, BWT, FWT, and LA are given as follow:  

\begin{equation}
{\text{Average Accuracy:    ACC}} = \frac{1}
{T}\sum\limits_{i = 1}^T {a_{T,i} } {\text{      }}
\label{13}
\end{equation}

\begin{equation}
{\text{Forgetting Measure:    FM}} = \frac{1}
{{T - 1}}\sum\limits_{j = 1}^{T - 1} {\mathop {\max }\limits_{l < T} a_{l,j}  - a_{T,j} } 
\label{14}
\end{equation}

\begin{equation}
{\text{Backward Transfer:    BWT}} = \frac{1}
{{T - 1}}\sum\limits_{i = 1}^{T - 1} {a_{T,i}  - } a_{i,i} {\text{  }}
\label{15}
\end{equation}

\begin{equation}
{\text{Forward Transfer:    FWT}} = \frac{1}
{{T - 1}}\sum\limits_{i = 1}^{T - 1} {a_{i - 1,i}  - \,} \bar b_i {\text{  }}
\label{16}
\end{equation}

\begin{equation}
{\text{Learning Accuracy:    LA}} = \frac{1}
{T}\sum\limits_{i = 1}^T {a_{i,i} } {\text{  }}
\label{17}
\end{equation}

The above five metrics evaluate the performance in the online continual learning environment from different perspectives. The average accuracy denotes the classification performance on all of the tasks when the algorithm finishes training process on the last task $T$. Forgetting is used to evaluate the model’s ability to maintain previous knowledge when the new knowledge is obtained, that means, the smaller this metric, the less forgetting. Next, BWT and FWT are used to measure the ability of knowledge transfer. Based on this, if two models enjoy similar average accuracies, we can further observe the transferability ability of the models. The larger BWF and FWF, the better the model. Lastly, learning accuracy is applied to measure the model’s ability about knowledge transfer, in other words, the knowledge transfer is accounted for the influence that the acquired knowledge is used to improve the performance of learning of latter tasks.

\subsection{Implementation Details}

We follow the implementation details in the literatures
\cite{44lopez2017gradient,45pham2020bilevel}, in all of our experiments. Particularly, all hyperparameters are optimized using grid-search. Moreover, in an online setting, learning is a “single pass through data” which means the agent only observes every training data once. The model for each example seeing only once could be deemed closer to real-world continual learning scenarios which is a very compelling setting. Thus, in order to simulate this scenario, we run all methods with one epoch for every task. For all datasets, we normalize the images and no other data augmentation is applied in this work. For a fair comparison, all methods use the same neural network architecture because they have the same number of parameters. In all of our experiments, we used a reduced ResNet18 
\cite{70he2016deep} as the baseline architecture, and details of network architecture are illustrated in the following Fig.4. For the training hyperparameters of experiments, we use SGD as our optimizer and the learning rates are set to 0.1, 0.1, 0.3, and 0.03 on Split CIFAR10, Split SVHN, Split CIFAR100 and Split TinyImagenet200 respectively. The batch size is set to 10 for all methods. Here, for memory-based methods used in this paper (Such as GEM, ER, DER and FTML), we keep the memory size consistent for every task. The size of episodic memory for every task is set to 65, 25, 65 and 128 on Split CIFAR10, Split SVHN, Split CIFAR100 and Split Tiny-ImageNet respectively. Lastly, for every baseline, the experiments are conducted five times, and the average ACC, FM, BWT, FWT, and LA are reported in this work.

\subsection{Baselines}

We extensively compare our MMKDDA method with several state-of-the-art baselines on various benchmark datasets, including \textbf{Finetune
\cite{44lopez2017gradient}}, \textbf{EWC
\cite{10kirkpatrick2017overcoming}}: Elastic Weight Consolidation, \textbf{LwF
\cite{20li2016learning}}: Learning without Forgetting, P\&C
\cite{22schwarz2018progress}: Progress and Compress, \textbf{ICARL
\cite{71rebuffi2017icarl}}: Incremental Classifier and Representation Learning, \textbf{GEM
\cite{44lopez2017gradient}}: Gradient Episodic Memory, \textbf{AGEM
\cite{72chaudhry2018efficient}}: Average Gradient Episodic Memory, \textbf{ER
\cite{26chaudhry2019continual}}: Experience Replay, \textbf{FTML
\cite{28finn2019online}}: Follow the Meta Leader. DER
\cite{47buzzega2020dark}: Dark Experience Replay. It is worth noting that MMKDDA-base denotes the method trained without a multi-scale knowledge distillation module, augmentation strategy, and the meta-learning update based on the number of tasks seen previously.

\subsection{Results and Analysis}

In this subsection, we conduct several groups of experiments to evaluate our MMKDDA algorithm in terms of ACC, FM, BWT, FWT, and LA. In the first series of experiments, we compare the performance of our MMKDDA algorithm to the state-of-the-art approaches in continual learning literature on different benchmark datasets and the average experimental results are reported by running the experiments five times. After that, we also perform the ablation studies to further analyze the contribution of each component for our proposed MMKDDA. The specific experimental results are shown in follow subsections.

\begin{figure}[H]
	\centering
	\includegraphics[width=\textwidth]{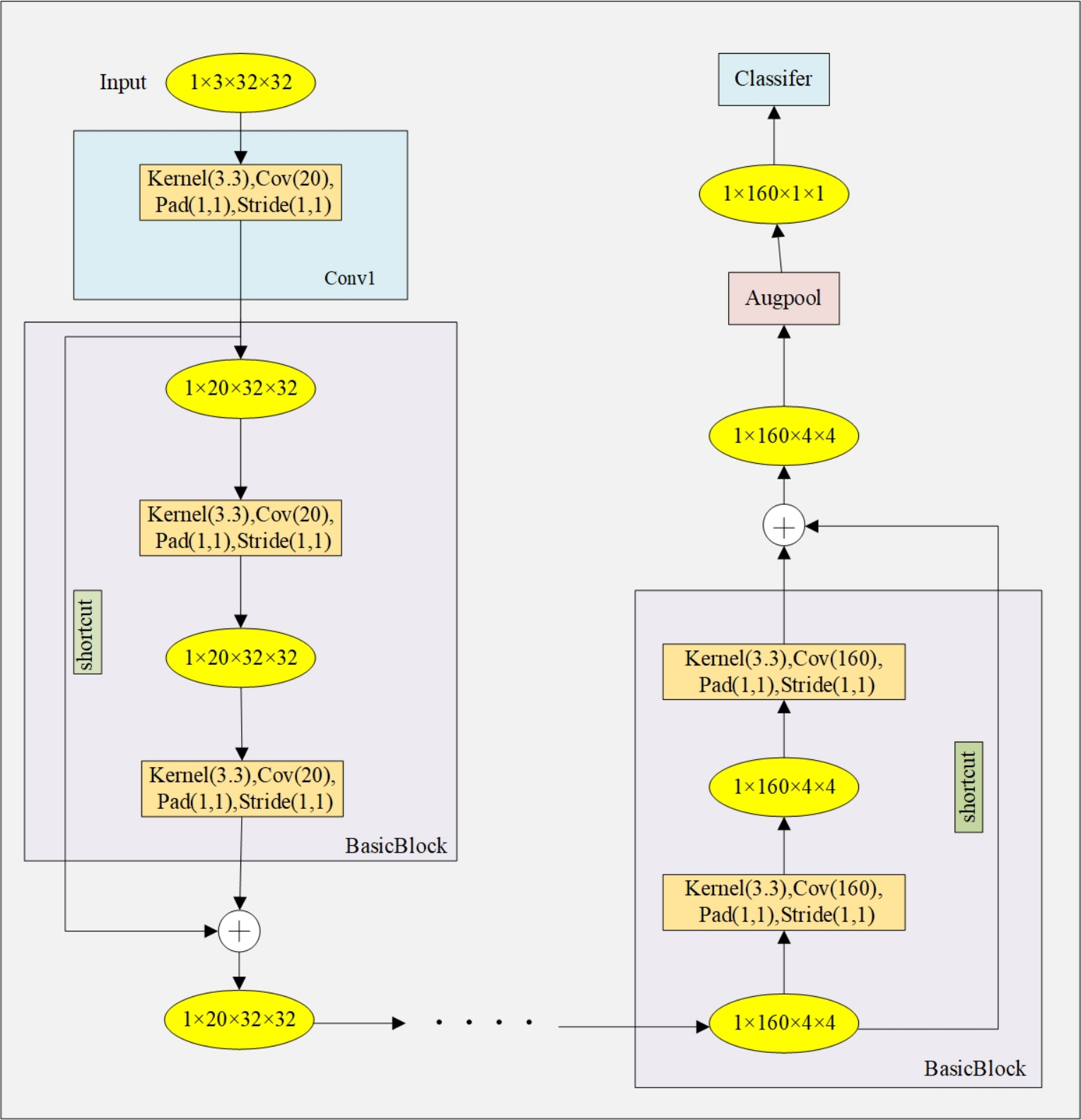}
	\caption{Details of model architectures used in this work on CIFAR dataset}
	\label{fig4}
\end{figure}

\subsubsection{Performance for MMKDDA on different datasets}

\textbf{1) Different Evaluation Metrics on various datasets}
 
Table 1 shows the numerical results of different baselines on five continual learning benchmarks: Split Cifar10(5T), Split SVHN(5T), Split Cifar100(20T), and Split Tiny-ImageNet(40T). Furthermore, the final percentage stack bar charts about the ACC and FM on different datasets are shown in Fig.5 and by percentage stack bar chart, we can observe the performance of various algorithms more clearly. Across all approaches, the value of ACC is high, indicating that this method exhibits better overall performance. The value of LA reflects the ability to learn new knowledge or adapting to a new task. Furthermore, FM value can measure the forgetting of old knowledge, and more specifically, this value is higher, catastrophic forgetting is more prominent. Lastly, knowledge transfer is measured by BWT and FWT.

From the experimental results, firstly we obviously observe that our base approach, even without a multi-scale knowledge distillation module, augmentation strategy, and the meta-learning update based on the number of tasks seen previously, can achieve competitive results in terms of almost all metrics considered in this paper. MMKDDA can further improve the performance of the algorithm. Our proposed MMKDDA algorithm outperforms all baselines on all datasets used in our experiments in terms of ACC. Then, we also find that traditional continual learning algorithms, such as finetune, LwF, and EWC, P\&C, exhibit relatively poor performance in an online setting on all datasets. Besides, we find that these methods based on memory strategy outperform the traditional methods on various evaluation metrics since the methods of these categories can alleviate catastrophic forgetting by episodic memory. For example, in terms of ACC, we observe that these approaches via memory strategy consistently provide an improvement by a large margin on all datasets. Besides, memory-based methods, such as GEM, AGEM, and ER, can alleviate catastrophic to a certain extent due to the neglect of knowledge transfer. Additionally, while the FTML algorithm can achieve favorable transferring ability with high LA values, this method cannot make a great balance between stability and plasticity, which results in lower overall performances compared to our MMKDDA. Compare to DER, our method also provides a further improvement.

\begin{table}[H]
  \centering
  \caption{The numerical results of different baselines on different datasets}
    \begin{tabular}{cccccc}
    \toprule
    \multirow{2}[2]{*}{Methods} & \multicolumn{5}{c}{Split Cifar10(5T)} \\
    \multicolumn{1}{c}{} & LA    & ACC   & BWT   & FWT   & FM \\
    \midrule
    Finetune & 88.24(1.91) & 70.49(6.08) & -22.19(5.42) & -0.94(1.47) & 26.15(8.16) \\
    EWC   & 87.94(3.38) & 73.98(4.96) & -17.45(4.37) & -1.28(5.29) & 18.73(5.10) \\
    LWF   & 88.75(1.86) & 75.74(2.16) & -16.27(3.08) & 0.39(3.89) & 16.27(3.08) \\
    P\&C  & 88.79(0.55) & 80.29(1.07) & -11.88(1.84) & 0.96(1.15) & 11.88(1.84) \\
    ICARL & 90.02(0.36) & 80.57(2.49) & -11.81(3.13) & 0.00(0.00) & 11.81(3.13) \\
    GEM   & 89.40(0.93) & 84.91(1.21) & -5.61(2.38) & 1.40(1.98) & 5.85(2.08) \\
    AGEM  & 88.70(1.85) & 84.42(2.20) & -5.35(2.49) & 0.17(3.02) & 6.58(2.22) \\
    ER    & 89.46(1.67) & 86.09(0.85) & -4.21(1.93) & 0.21(2.08) & 4.61(1.22) \\
    FTML  & 90.99(0.95) & 85.93(1.09) & -6.33(2.25) & 0.50(1.34) & 6.41(2.09) \\
    DER   & 89.99(053) & 86.26(1.18) & -4.66(1.17) & 2.30(2.47) & 4.82(1.15) \\
    Our base & 90.79(0.54) & 86.50(0.82) & -5.37(0.89) & 1.67(1.56) & 5.49(0.99) \\
    MMKDDA & 90.32(0.46) & 88.89(0.21) & -1.79(0.60) & 1.23(3.62) & 2.18(0.27) \\
    \bottomrule
    \end{tabular}%
  \label{tab:addlabel}%
\end{table}%

\begin{table}[H]
  \centering

    \begin{tabular}{cccccc}
    \toprule
    \multirow{2}[2]{*}{Methods} & \multicolumn{5}{c}{Split SVHN(5T)} \\
    \multicolumn{1}{c}{} & LA    & ACC   & BWT   & FWT   & FM \\
    \midrule
    Finetune & 96.07(0.12) & 75.87(7.18) & -25.25(9.01) & -0.44(3.83) & 25.25(9.01) \\
    EWC   & 94.39(1.60) & 82.65(1.79) & -14.66(4.10) & -3.21(2.16) & 14.66(4.10) \\
    LWF   & 93.97(1.11) & 83.27(1.96) & -13.37(3.65) & 3.56(5.32) & 13.37(3.65) \\
    P\&C  & 96.39(1.88) & 86.59(1.87) & -9.75(0.71) & 0.18(3.00) & 9.75(0.71) \\
    ICARL & 95.39(1.04) & 86.89(2.04) & -10.62(3.54) & 0.00(0.00) & 10.62(3.54) \\
    GEM   & 95.72(1.02) & 90.00(1.59) & -7.14(3.08) & 3.08(4.36) & 7.14(3.08) \\
    AGEM  & 95.26(0.66) & 90.84(1.83) & -5.54(2.13) & 2.71(6.34) & 5.54(2.13) \\
    ER    & 96.43(0.41) & 91.57(1.37) & -6.06(1.82) & 1.59(4.99) & 6.08(1.82) \\
    FTML  & 96.68(0.21) & 93.30(0.63) & -4.22(1.03) & -0.15(2.80) & 4.22(1.03) \\
    DER   & 95.88(0.60) & 93.02(0.91) & -3.57(1.52) & -2.62(9.57) & 3.67(1.49) \\
    Our base & 96.33(0.39) & 92.79(0.53) & -4.42(0.71) & 1.14(3.71) & 4.42(0.71) \\
    MMKDDA & 95.80(0.64) & 95.03(0.59) & -0.97(0.37) & 3.87(3.50) & 1.02(0.31) \\
    \midrule
    \multicolumn{1}{c}{} & \multicolumn{1}{r}{} & \multicolumn{1}{r}{} & \multicolumn{1}{r}{} & \multicolumn{1}{r}{} & \multicolumn{1}{r}{} \\
    \midrule
    \multirow{2}[2]{*}{Methods} & \multicolumn{5}{c}{Split Cifar100(20T)} \\
    \multicolumn{1}{c}{} & LA    & ACC   & BWT   & FWT   & FM \\
    \midrule
    Finetune & 66.01(0.41) & 34.93(1.27) & -32.71(1.22) & 1.71(0.75) & 92.75(1.23) \\
    EWC   & 64.38(0.92) & 38.24(4.40) & -26.47(4.91) & 0.485(1.70) & 26.86(4.73) \\
    LWF   & 64.25(0.77) & 42.24(1.52) & -22.11(1.16) & -0.94(0.88) & 22.40(1.28) \\
    P\&C  & 67.29(1.69) & 48.39(3.55) & -19.05(5.05) & 0.35(1.16) & 19.70(4.28) \\
    ICARL & 66.01(1.99) & 46.48(0.61) & -18.45(2.30) & 0.00(0.00) & 18.77(2.12) \\
    GEM   & 67.99(0.71) & 62.63(0.74) & -5.65(0.87) & -0.19(1.38) & 7.08(0.52) \\
    AGEM  & 67.56(1.07) & 55.59(1.82) & -12.61(1.33) & 0.07(1.50) & 13.27(1.03) \\
    ER    & 68.73(1.07) & 63.04(1.99) & -5.99(1.31) & 0.23(1.03) & 5.83(2.76) \\
    FTML  & 70.23(0.48) & 62.72(1.50) & -7.9(1.46) & 0.38(1.11) & 8.50(1.44) \\
    DER   & 69.53(1.27) & 65.10(0.83) & -4.66(0.84) & -0.85(1.56) & 6.11(0.39) \\
    Our base & 68.73(1.22) & 64.46(0.67) & -4.50(0.78) & 1.04(1.01) & 5.59(0.75) \\
    MMKDDA & 69.29(0.71) & 67.81(0.64) & -1.56(0.57) & -0.19(1.02) & 2.98(0.58) \\
    \bottomrule
    \end{tabular}%
\end{table}%

\begin{table}[H]
  \centering
    \begin{tabular}{cccccc}
    \toprule
    \multirow{2}[2]{*}{Methods} & \multicolumn{5}{c}{Split Tiny-ImageNet(40T)} \\
    \multicolumn{1}{c}{} & LA    & ACC   & BW    & FW    & FM \\
    \midrule
    Finetune & 64.82(0.62) & 34.10(3.57) & -31.51(3.87) & -0.16(0.34) & 31.54(3.84) \\
    EWC   & 65.13(1.26) & 35.09(1.70) & -30.81(1.25) & -1.14(0.75) & 30.88(1.20) \\
    LWF   & 64.77(0.29) & 36.41(0.38) & -29.09(0.49) & -0.15(0.84) & 29.12(0.46) \\
    P\&C  & 66.57(0.64) & 45.69(9.13) & -22.43(9.43) & 0.06(1.34) & 22.45(9.41) \\
    ICARL & 66.19(0.68) & 43.87(0.40) & -22.89(0.79) & 0.00(0.00) & 23.04(0.73) \\
    GEM   & 67.00(0.81) & 63.24(0.99) & -3.85(0.25) & -0.11(0.81) & 6.46(0.55) \\
    AGEM  & 66.75(0.48) & 60.37(1.01) & -6.55(1.35) & 0.34(1.32) & 7.92(1.07) \\
    ER    & 68.48(0.43) & 67.44(0.24) & -1.06(0.62) & -0.36(0.75) & 3.65(0.46) \\
    FTML  & 72.69(0.64) & 63.84(0.76) & -9.07(0.60) & -0.06(0.57) & 9.38(0.67) \\
    DER   & 69.29(0.99) & 67.50(0.55) & -1.83(0.64) & 0.53(0.69) & 4.58(0.22) \\
    Our base & 71.74(0.74) & 65.79(0.80) & -6.10(0.48) & -0.06(1.08) & 6.91(0.40) \\
    MMKDDA & 72.10(0.74) & 68.78(0.98) & -3.39(0.55) & -0.36(0.47) & 4.67(0.44) \\
    \bottomrule
    \end{tabular}%
\end{table}%

\begin{figure}[H]
	\centering
	\subfigure[]{
		\includegraphics[width=0.9\textwidth]{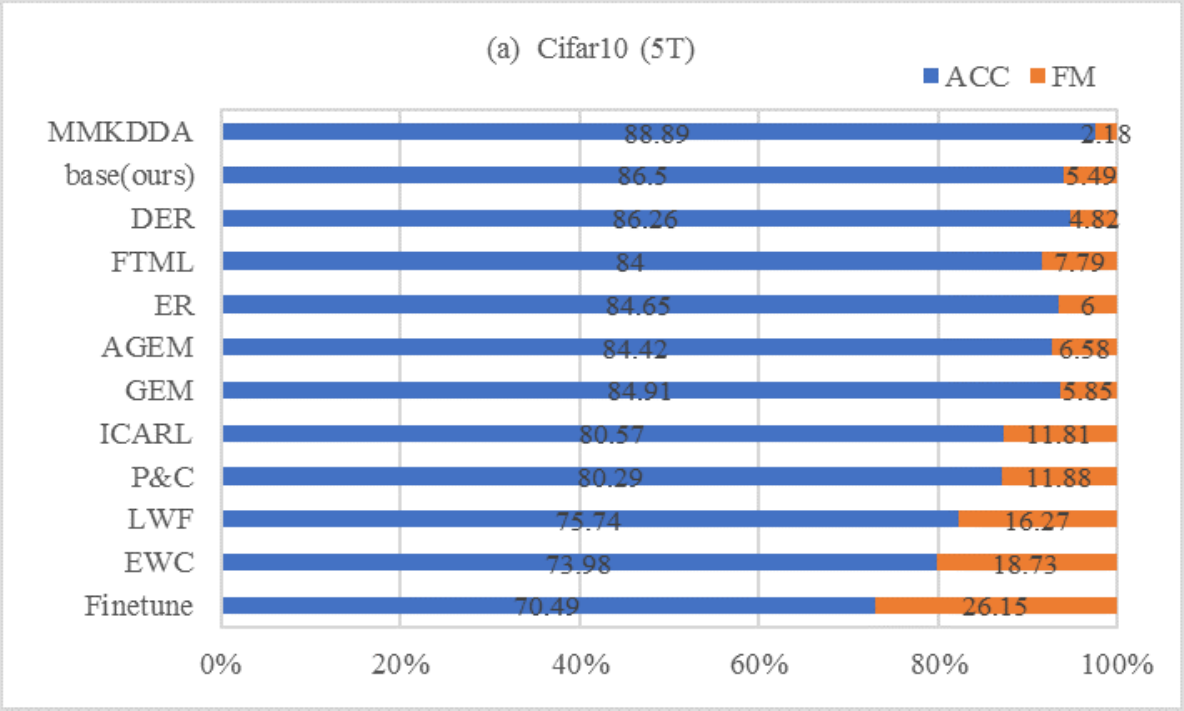}
	}
\end{figure}
\addtocounter{figure}{-1}
\begin{figure}[H]
\addtocounter{figure}{1}
	\centering
	\subfigure[]{
		\includegraphics[width=0.9\textwidth]{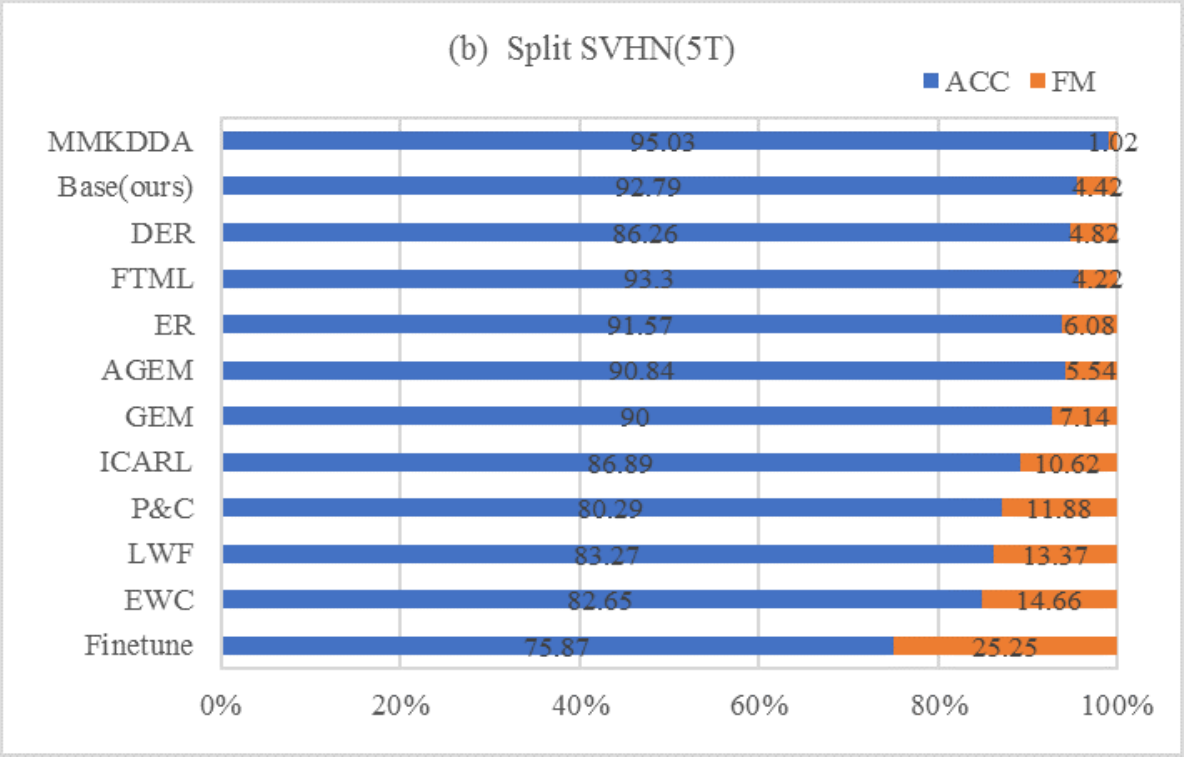}
	}
	\subfigure[]{
		\includegraphics[width=0.9\textwidth]{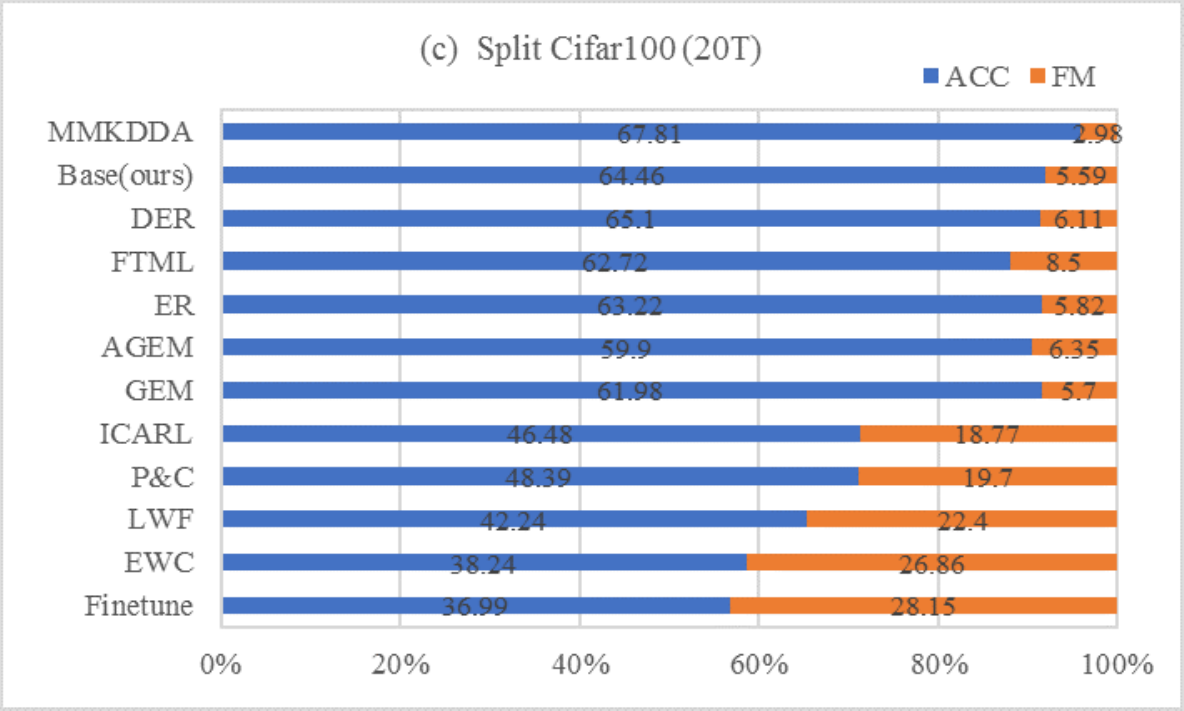}
	}
\end{figure}
\addtocounter{figure}{-1}
\begin{figure}[H]
\addtocounter{figure}{1}
	\centering
	\subfigure[]{
		\includegraphics[width=0.9\textwidth]{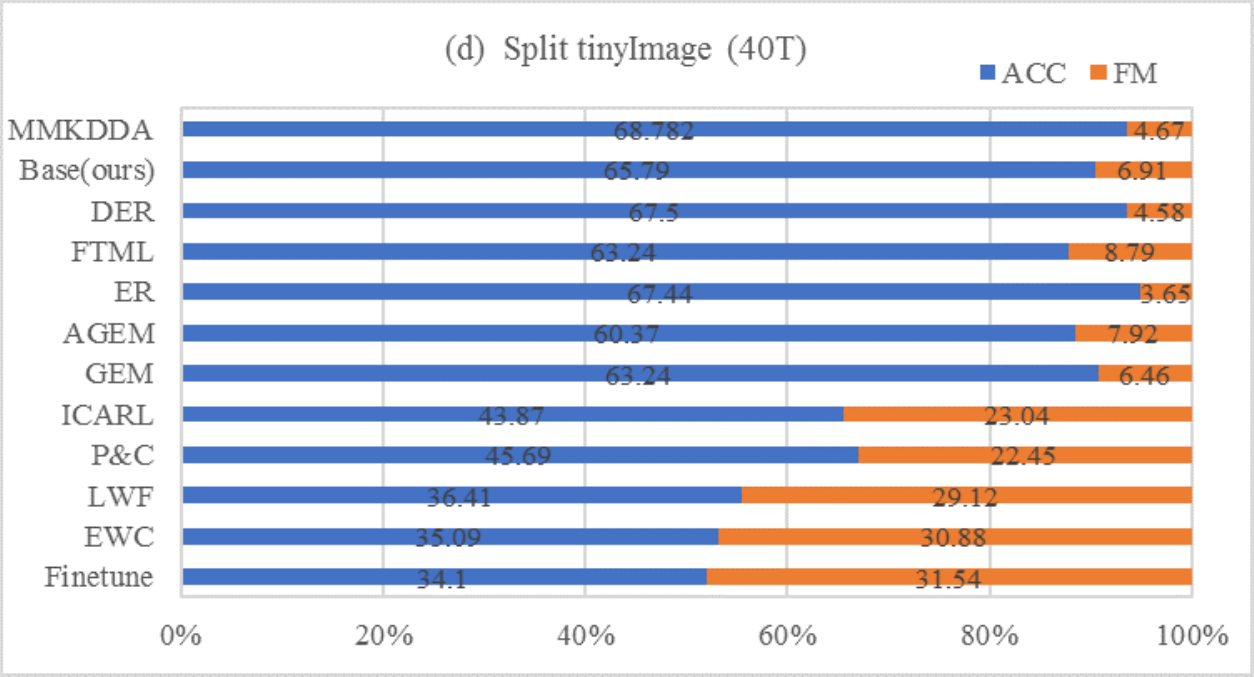}
	}
     \caption{The final percentage stack bar chart about the ACC and FM}
     \label{fig5}
\end{figure}

\textbf{2) The evolution of average accuracy as the number of tasks increases}

Fig.6 shows that the evolution of ACCs as a new task is trained in different datasets while the final accuracy for each task is shown in Fig.7. From the experimental results, Fig.6 shows that when the number of tasks is small, all approaches enjoy small differences, and as more tasks are observed, our proposed MMKDDA achieves superior performance than other baselines on all datasets. That means, our MMKDDA can effectively mitigate catastrophic forgetting in the context of online continual learning. Then, we also find that the performance of traditional methods declines dramatically as the number of tasks rises, which means, these approaches have limited ability to overcome catastrophic forgetting. Furthermore, those methods based on memory strategy show that the downward trend is relatively slow, which indicates that these methods with memory strategy can reduce forgetting to a certain extend. Fig.7 plots the final accuracy for each task when the whole tasks have been trained, and then we can further observe the accuracy for each task. Notably, our MMKDDA exhibits superior performance than other baselines, which also validates that our MMKDDA is a promising approach.

\begin{figure}[H]
	\centering
	\subfigure[]{
		\includegraphics[width=0.9\textwidth]{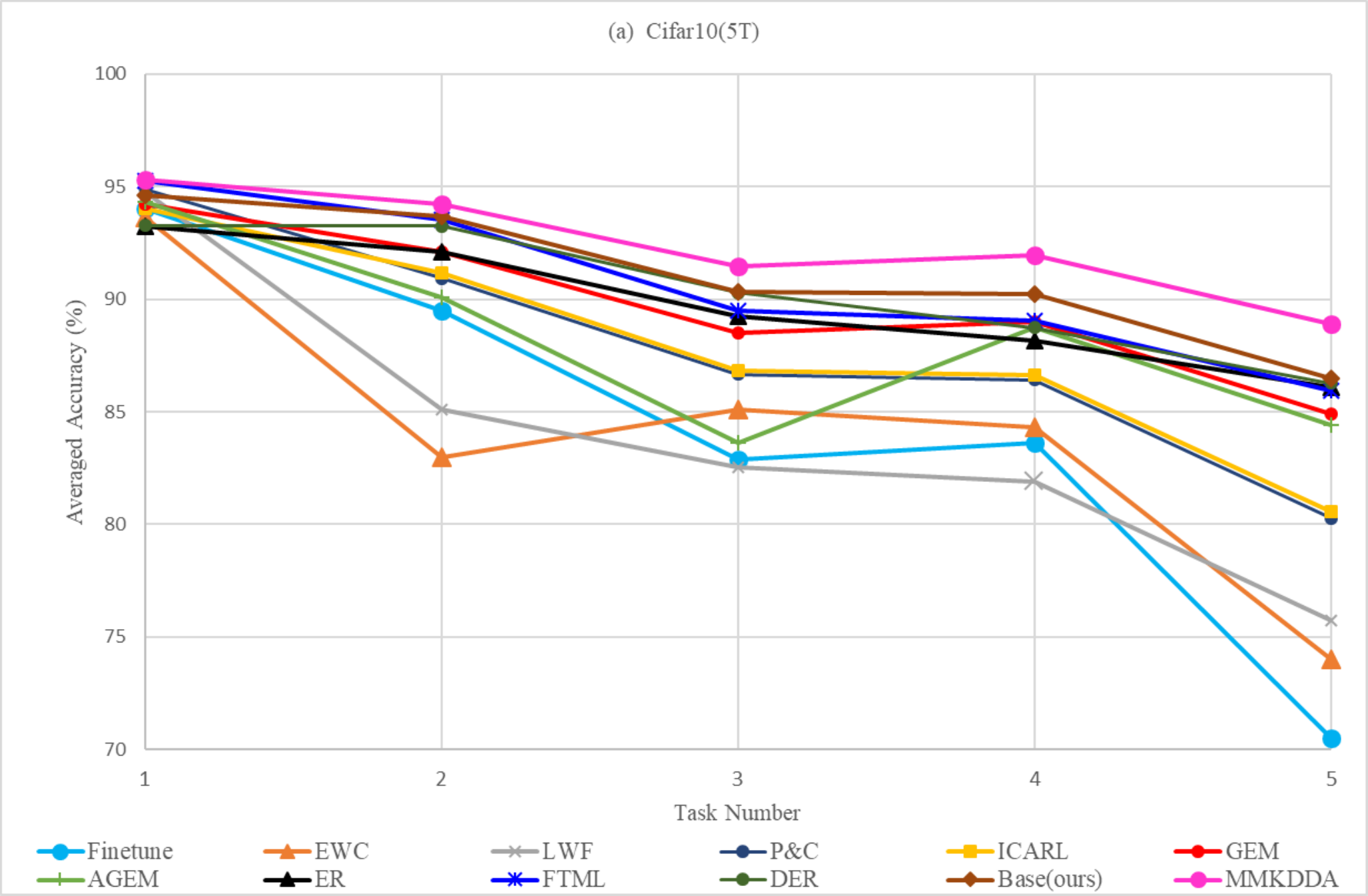}
	}
     \\
     \subfigure[]{
		\includegraphics[width=0.9\textwidth]{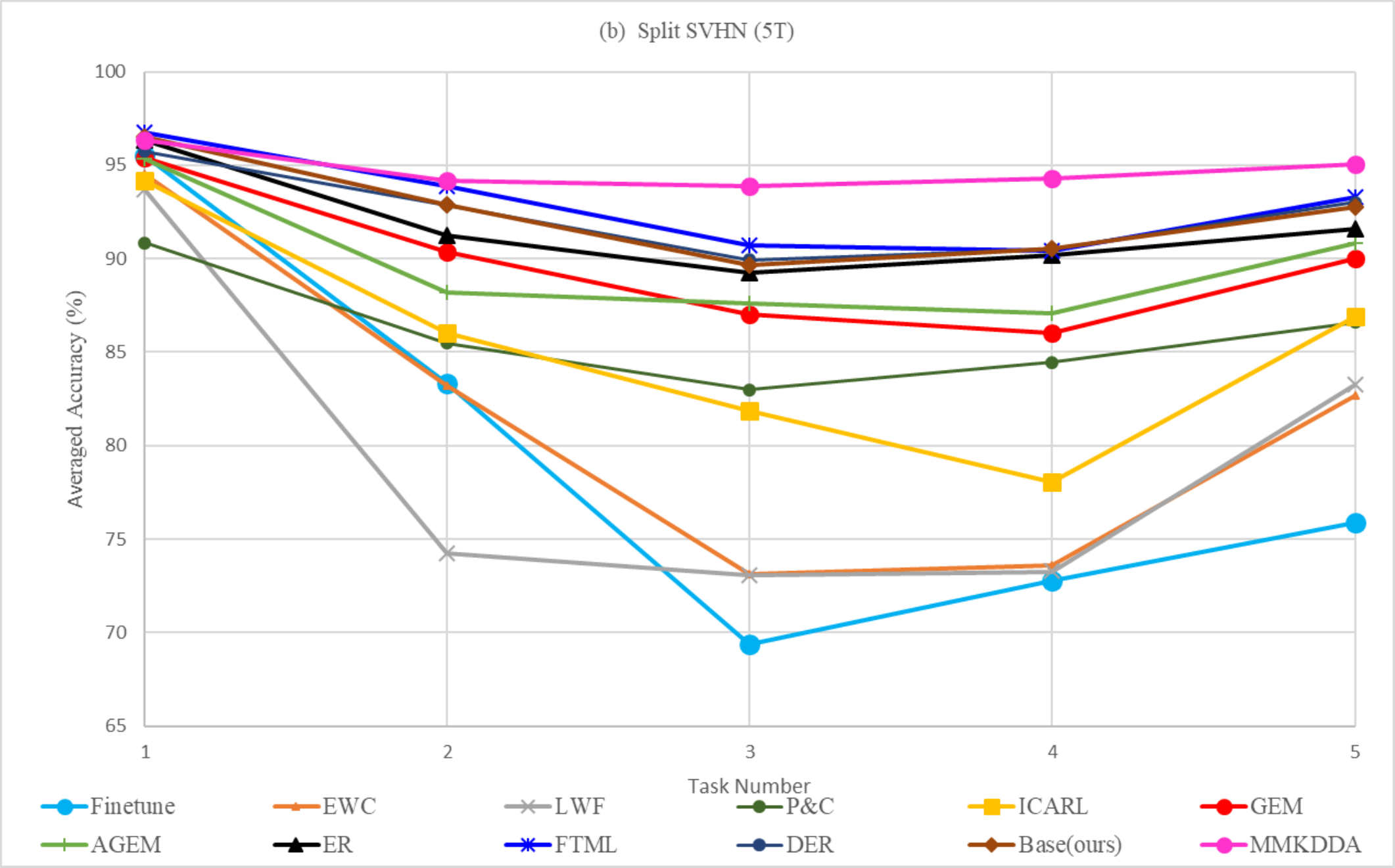}
	}
\end{figure}
\addtocounter{figure}{-1}
\begin{figure}[H]
\addtocounter{figure}{1}
	\centering
	\subfigure[]{
		\includegraphics[width=0.9\textwidth]{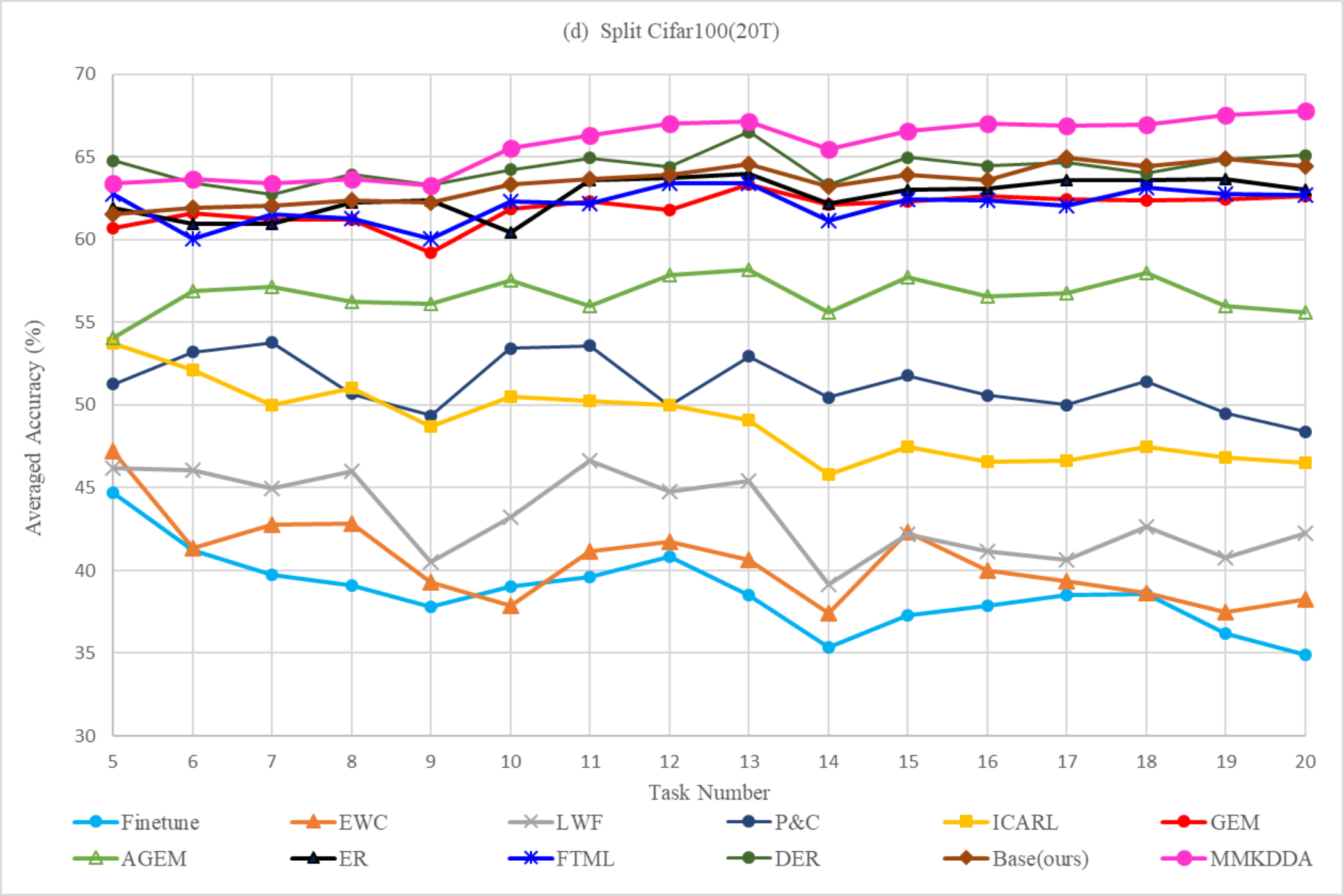}
	}
	\subfigure[]{
		\includegraphics[width=0.9\textwidth]{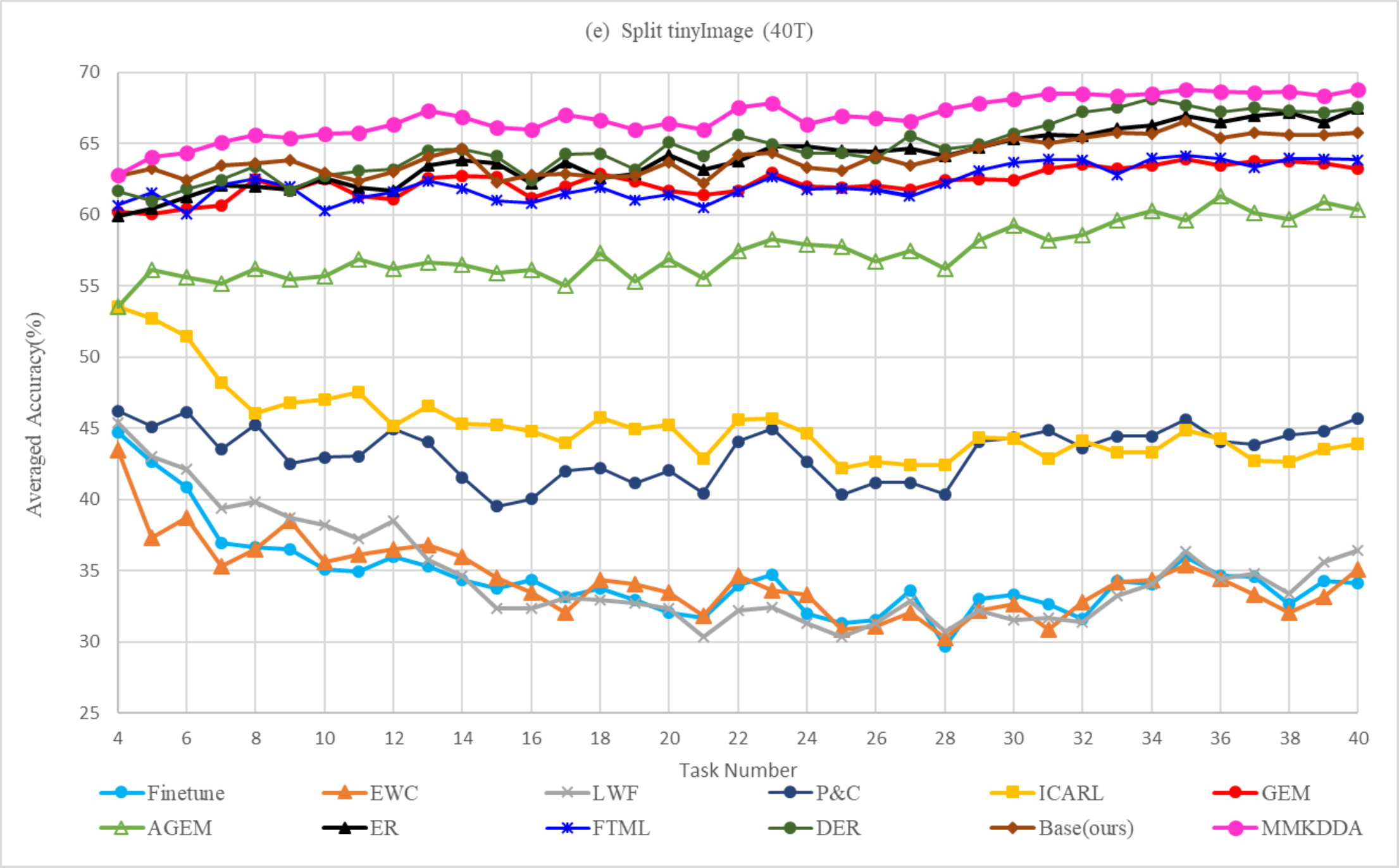}
	}
     \caption{Evolution curve of average accuracy as new tasks are learned}
     \label{fig6}
\end{figure}

\begin{figure}[H]
	\centering
	\subfigure[]{
		\includegraphics[width=0.9\textwidth]{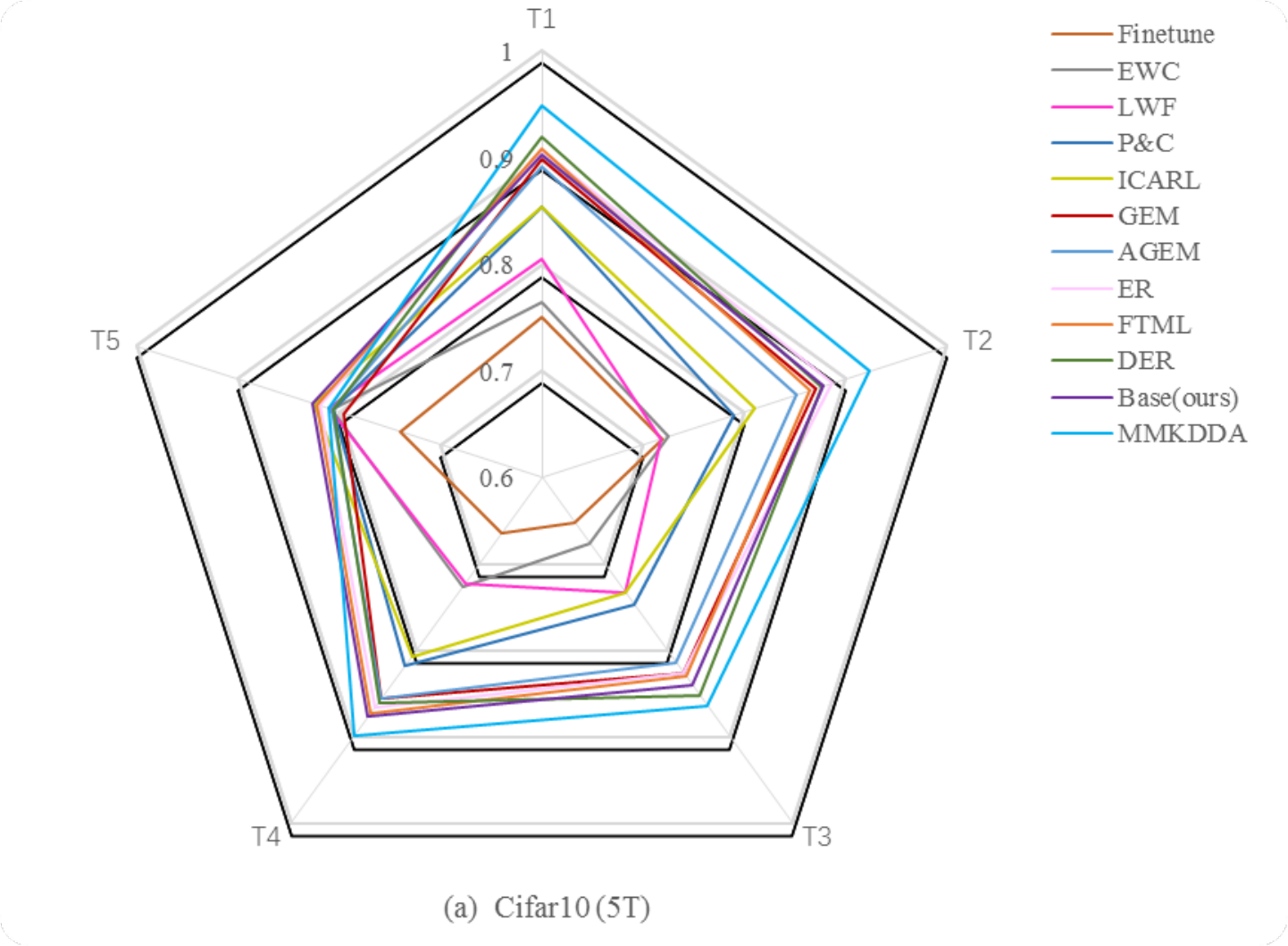}
	}
     \\
     \subfigure[]{
		\includegraphics[width=0.9\textwidth]{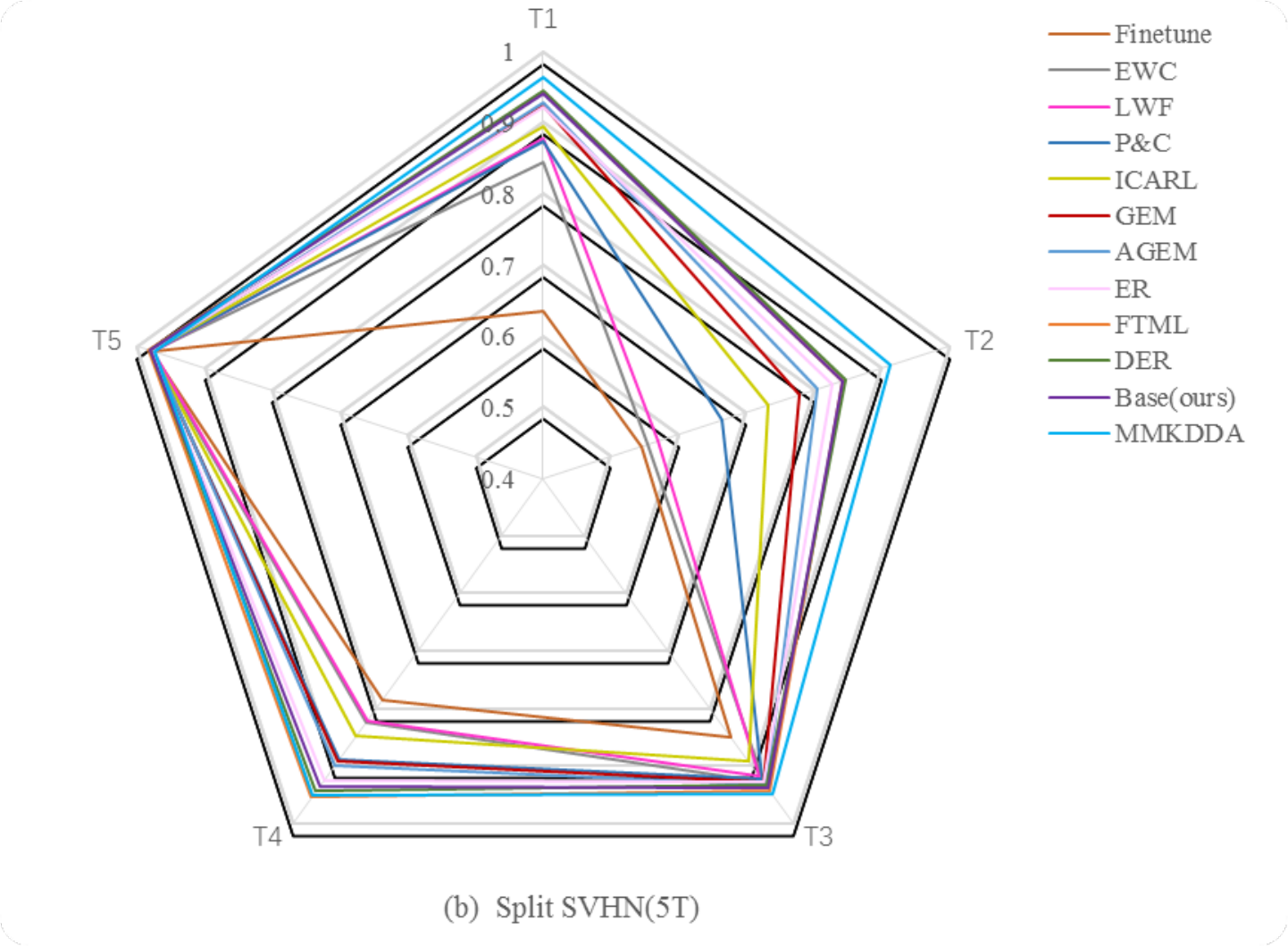}
	}
\end{figure}
\addtocounter{figure}{-1}
\begin{figure}[H]
\addtocounter{figure}{1}
	\centering
	\subfigure[]{
		\includegraphics[width=0.8\textwidth]{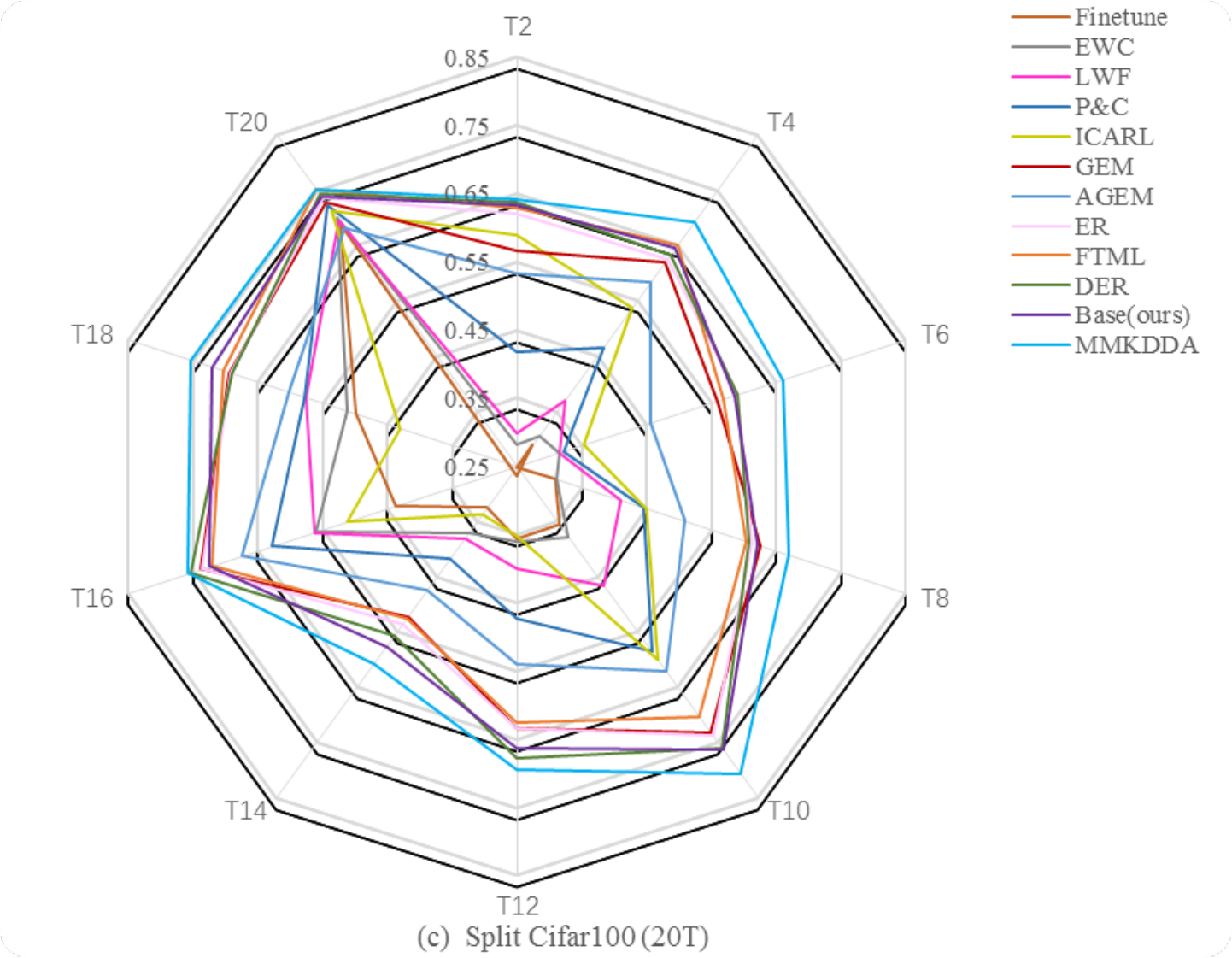}
	}
	\subfigure[]{
		\includegraphics[width=0.8\textwidth]{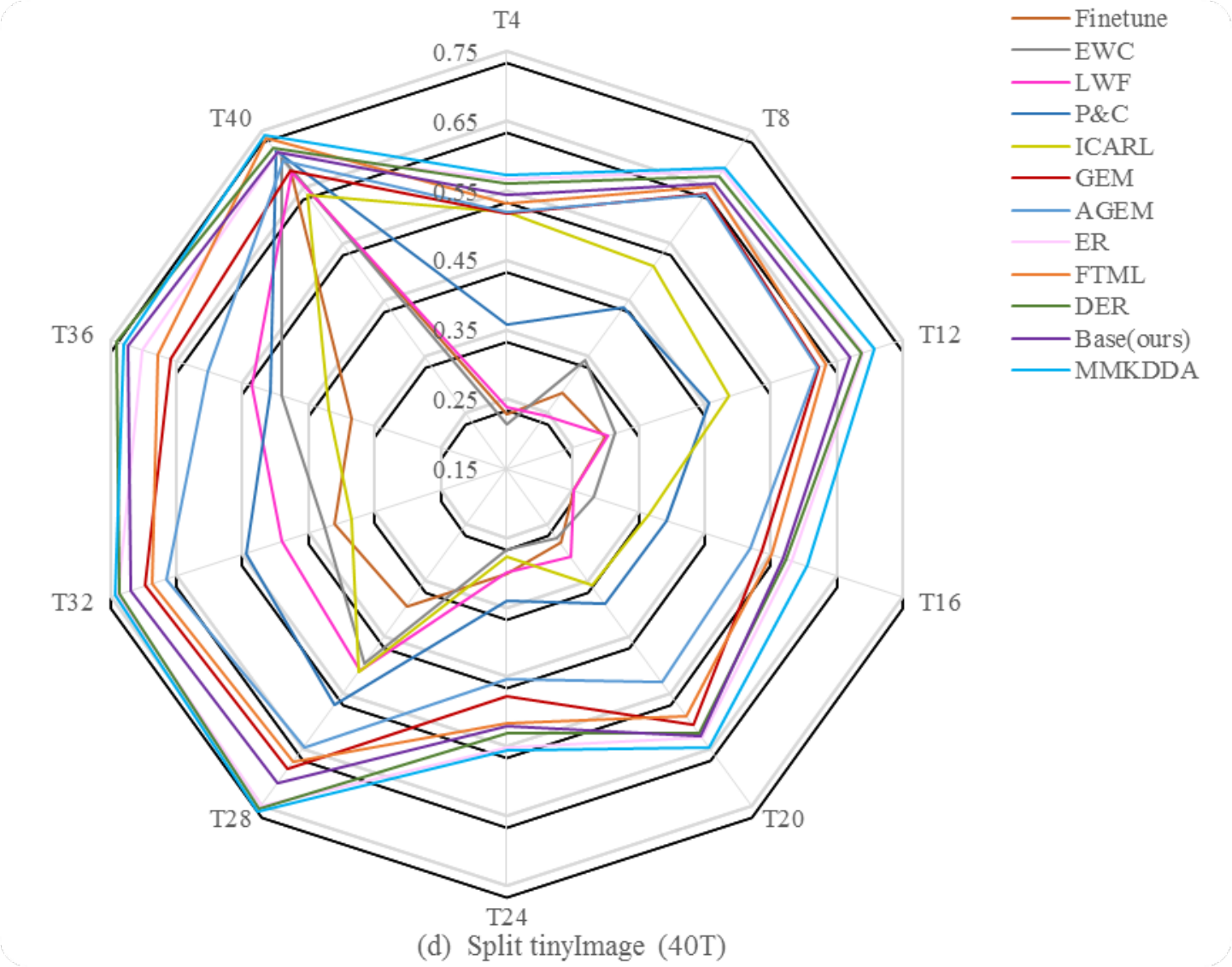}
	}
     \caption{The final accuracies on each task on different datasets}
     \label{fig7}
\end{figure}

\subsubsection{Ablation studies on different datasets}

\textbf{1) Different evaluation metrics on various datasets}

In this subsection, in order to further understand the contribution of each component for our MMKDDA, we perform various ablation studies on Split SVHN, Split CIFAR100, and Split TinyImagenet200 datasets respectively. Particularly, we are interested in how (1) multi-scale knowledge distillation (MKD), (2) data augmentation (DA), and (3) meta-learning update based on the number of tasks seen previously (ML) contribute to the base model. We conduct the ablation experiments with various combinations of these components and the results are provided in Table 2. Here, the provided metrics reflect different aspects of continual learning. The base model denotes the performance without MKD, DA and ML. In sum up, 1) MKD: The multi-scale knowledge distillation is added to constrain the evolution of long-range and short-range spatial relationships; 2) DA: The data augmentation is used to build the training data; 3) MKD+DA: The MKD and DA strategy are employed simultaneously; 4) MKD+DA + ML: based on MKD+DA, the meta-learning update based on the number of tasks seen previously is further added. Finally, we can analyze the effect of MKD and DA on the model via experiment 1) and experiment 2) respectively. Experiment 4) further shows the contribution of the ML. 

Now, we analyze how every component in MMKDDA affects its overall performance. In table 2, firstly, we can observe that MKD+DA + ML achieves the highest accuracies and the lowest forgetting on all datasets used in this work, which means this method provides the most favorable performance which is also found easily from Fig.8. More specifically, compared with base method, the MMKDDA gains 2.24\%, 3.35\%, and 1.77\% improvements on Split SVHN, Split CIFAR100, and Split Tiny-Imagenet datasets, respectively. Secondly, we can see that the base method achieves the lowest performance compared with other approaches. It is worth noting that when the multi-scale knowledge distillation and data augmentation strategy are employed respectively, the performance of the base method can be consistently improved, e.g. on Split CIFAR100, 1) +MKD, 2) +DA, and 3) MKD+DA respectively gains 1.27\%, 0.51\%, and 1.83\% on ACC. Besides, we also find that when the meta-learning update based on the number of tasks seen previously is employed, the performance can be further improved. The experimental results confirm our previous conjecture, and each component in our MMKDDA has its own contribution.

\begin{table}[H]
  \centering
  \caption{The numerical results of different ablation studies on different datasets}
    \begin{tabular}{ccccccc}
    \toprule
    \multicolumn{1}{c}{\multirow{2}[4]{*}{Method}} & \multirow{2}[4]{*}{MKD} & \multirow{2}[4]{*}{DA} & \multirow{2}[4]{*}{ML} & \multicolumn{3}{c}{Split Cifar100(20T)} \\
\cmidrule{5-7}          & \multicolumn{1}{c}{} & \multicolumn{1}{c}{} & \multicolumn{1}{c}{} & ACC   & LA    & FM \\
    \midrule
    \multicolumn{1}{c}{Base-model} & \multicolumn{1}{c}{} & \multicolumn{1}{c}{} & \multicolumn{1}{c}{} & 64.46(0.67) & 68.73(1.22) & 5.59(0.75) \\
    \midrule
    1     & \checkmark     & \multicolumn{1}{c}{} & \multicolumn{1}{c}{} & 65.73(1.07) & 67.83(0.63) & 3.75(0.81) \\
    2     & \multicolumn{1}{c}{} & \checkmark     & \multicolumn{1}{c}{} & 64.97(1.08) & 69.24(0.59) & 5.70(1.50) \\
    3     & \checkmark     & \checkmark     & \multicolumn{1}{c}{} & 66.29(0.87) & 68.99(0.57) & 4.23(1.10) \\
    4     & \checkmark     & \checkmark     & \checkmark     & \textbf{67.81(0.64)} & \textbf{69.29(0.71)} & \textbf{2.98(0.58)} \\
    \midrule
          & \multicolumn{1}{r}{} & \multicolumn{1}{r}{} & \multicolumn{1}{r}{} & \multicolumn{1}{r}{} & \multicolumn{1}{r}{} & \multicolumn{1}{r}{} \\
    \midrule
    \multicolumn{1}{c}{\multirow{2}[4]{*}{Method}} & \multirow{2}[4]{*}{MKD} & \multirow{2}[4]{*}{DA} & \multirow{2}[4]{*}{ML} & \multicolumn{3}{c}{Split SVHN (5T)} \\
\cmidrule{5-7}          & \multicolumn{1}{c}{} & \multicolumn{1}{c}{} & \multicolumn{1}{c}{} & ACC   & LA    & FM \\
    \midrule
    \multicolumn{1}{c}{Base-model} & \multicolumn{1}{c}{} & \multicolumn{1}{c}{} & \multicolumn{1}{c}{} & 92.79(0.53) & 96.33(0.39) & 4.42(0.71) \\
    \midrule
    1     & \checkmark     & \multicolumn{1}{c}{} & \multicolumn{1}{c}{} & 94.31(0.29) & 95.86(0.44) & 2.03(0.40) \\
    2     & \multicolumn{1}{c}{} & \checkmark     & \multicolumn{1}{c}{} & 93.25(0.51) & \textbf{96.64(0.37)} & 4.23(0.70) \\
    3     & \checkmark     & \checkmark     & \multicolumn{1}{c}{} & 94.89(0.57) & 96.40(0.26) & 1.92(0.82) \\
    4     & \checkmark     & \checkmark     & \checkmark     & \textbf{95.03(0.59)} & 95.80(0.64) & \textbf{1.02(0.31)} \\
    \midrule
          & \multicolumn{1}{r}{} & \multicolumn{1}{r}{} & \multicolumn{1}{r}{} & \multicolumn{1}{r}{} & \multicolumn{1}{r}{} & \multicolumn{1}{r}{} \\
    \midrule
    \multicolumn{1}{c}{\multirow{2}[4]{*}{Method}} & \multirow{2}[4]{*}{MKD} & \multirow{2}[4]{*}{DA} & \multirow{2}[4]{*}{ML} & \multicolumn{3}{c}{Split Tiny-Image-Net(40T)} \\
\cmidrule{5-7}          & \multicolumn{1}{c}{} & \multicolumn{1}{c}{} & \multicolumn{1}{c}{} & ACC   & LA    & FM \\
    \midrule
    \multicolumn{1}{c}{Base-model} & \multicolumn{1}{c}{} & \multicolumn{1}{c}{} & \multicolumn{1}{c}{} & 65.79(0.80) & 71.74(0.74) & 6.91(0.40) \\
    \midrule
    1     & \checkmark     & \multicolumn{1}{c}{} & \multicolumn{1}{c}{} & 67.31(0.54) & 71.13(0.52) & 5.29(0.43) \\
    2     & \multicolumn{1}{c}{} & \checkmark     & \multicolumn{1}{c}{} & 66.37(0.71) & 72.02(0.52) & 6.18(0.58) \\
    3     & \checkmark     & \checkmark     & \multicolumn{1}{c}{} & 67.34(0.38) & 71.69(0.44) & 5.68(0.28) \\
    4     & \checkmark     & \checkmark     & \checkmark     & \textbf{68.78(0.98)} & \textbf{72.10(0.74)} & \textbf{4.67(0.44)} \\
    \bottomrule
    \end{tabular}%
  \label{tab:addlabel}%
\end{table}%

\begin{figure}[H]
	\centering
	\subfigure[]{
		\includegraphics[width=\textwidth]{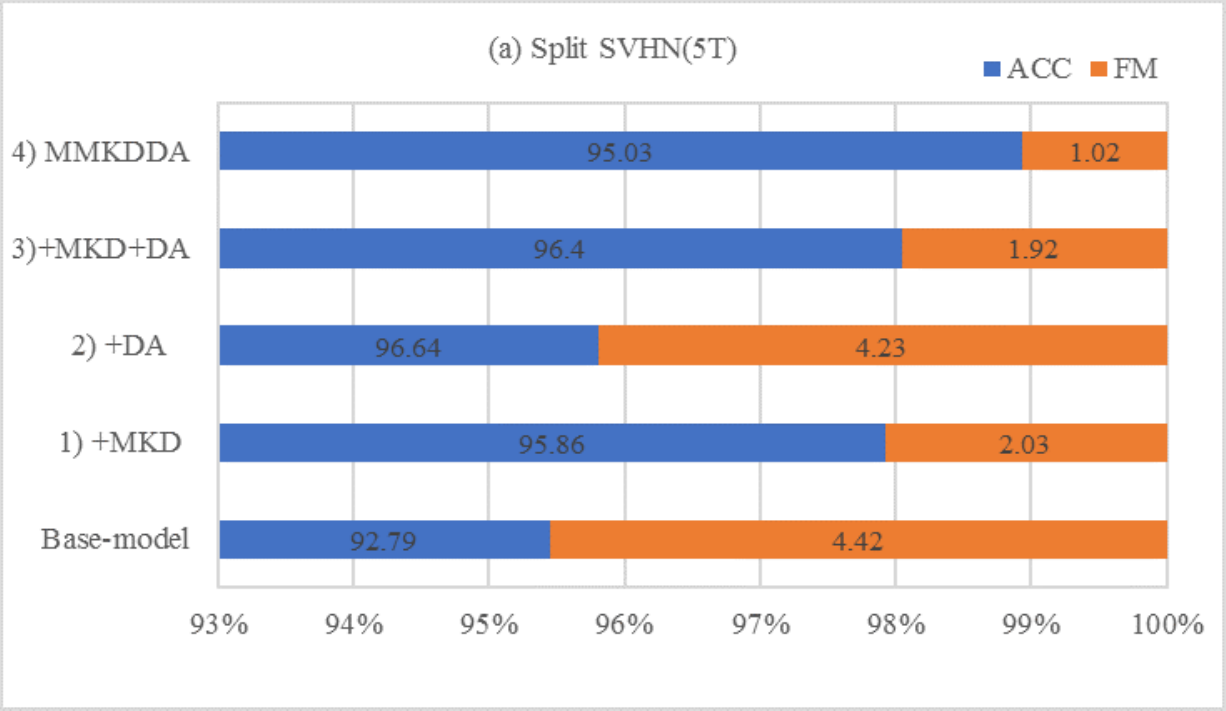}
	}
     \\
     \subfigure[]{
		\includegraphics[width=\textwidth]{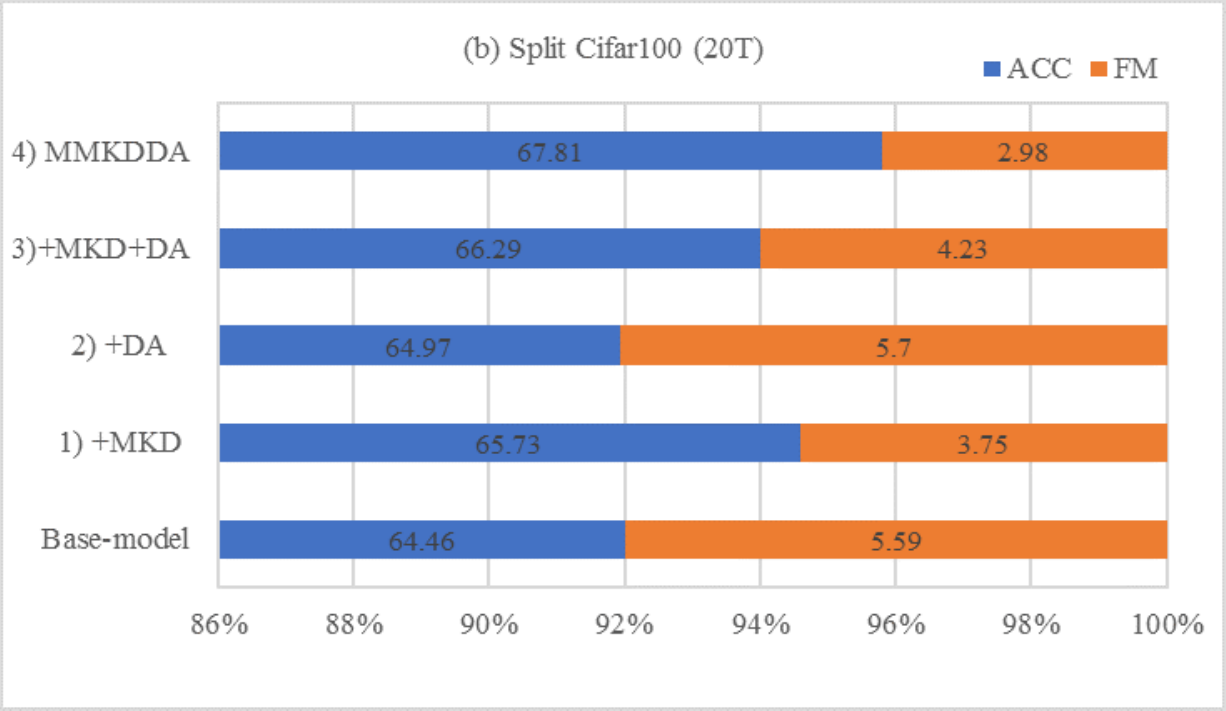}
	}
\end{figure}
\addtocounter{figure}{-1}
\begin{figure}[H]
\addtocounter{figure}{1}
	\centering
	\subfigure[]{
		\includegraphics[width=\textwidth]{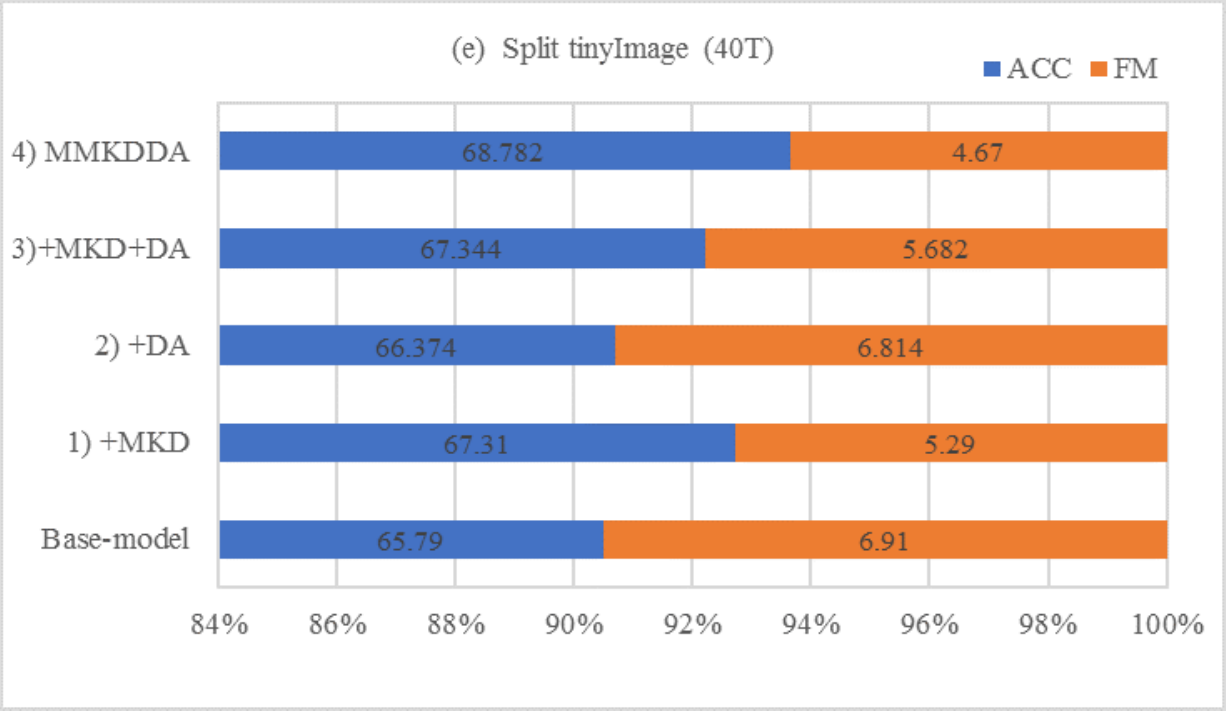}
	}
     \caption{The final percentage stack bar chart about the ACC and FM on ablation studies}
     \label{fig8}
\end{figure}

\textbf{2)The evolution curves of average accuracies as increasing the number of tasks}

Fig.8 plots the evolution curves of average accuracies as more tasks arrive and Fig.9 plots the final test accuracies at each task when the training is completed. The previous findings can be further confirmed by observing Fig.8 and Fig.9. More specifically, firstly, as more tasks come, 4) MKD+DA + ML achieves superior performance than other methods. Besides, compared with other methods, base approach suffers from relatively poor performance. Meanwhile, when different components are applied in our model, the performance of the algorithm is improved to some extent. For instance, when the multi-scale knowledge distillation (MKD) module is employed, the ability to overcome catastrophic forgetting is significantly enhanced. Furthermore, we can see that when the training process is completed, from Fig.9, we observe that our MMKDDA method outperforms other methods on almost every task. These results show that our proposed MMKDDA is a promised approach in the context of online continual learning.

\begin{figure}[H]
	\centering
	\subfigure[]{
		\includegraphics[width=\textwidth]{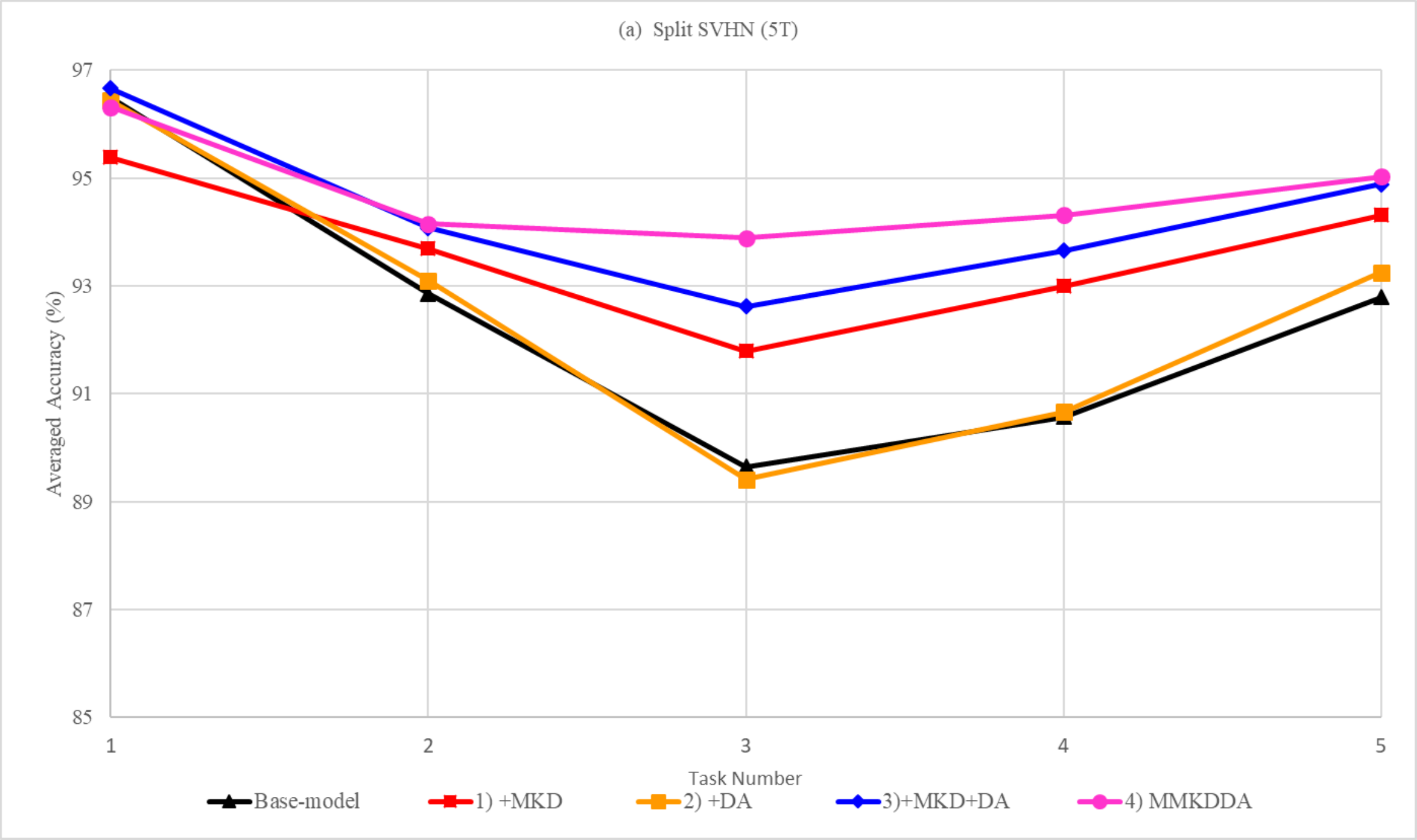}
	}
     \\
     \subfigure[]{
		\includegraphics[width=\textwidth]{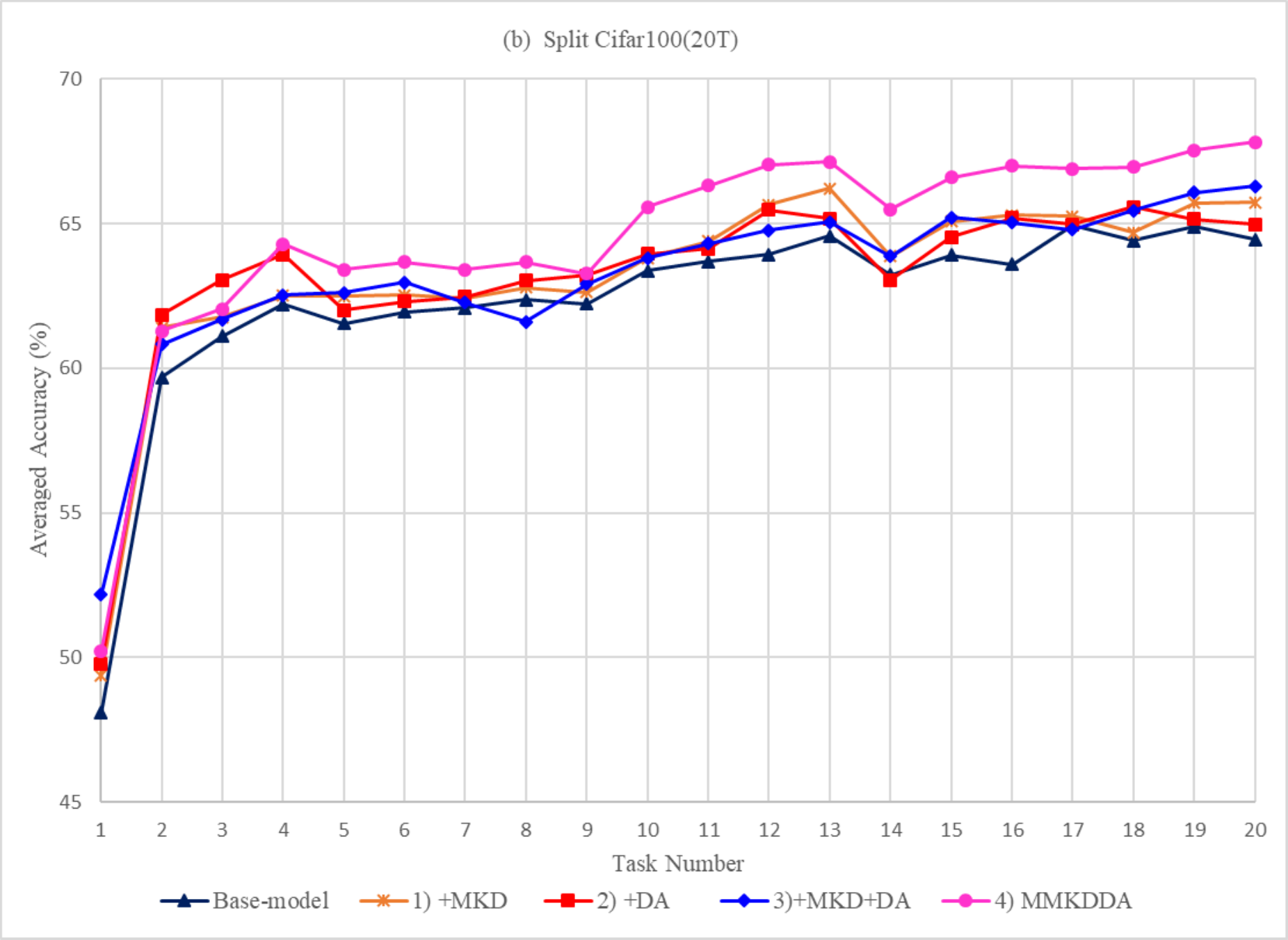}
	}
\end{figure}
\addtocounter{figure}{-1}
\begin{figure}[H]
\addtocounter{figure}{1}
	\centering
	\subfigure[]{
		\includegraphics[width=\textwidth]{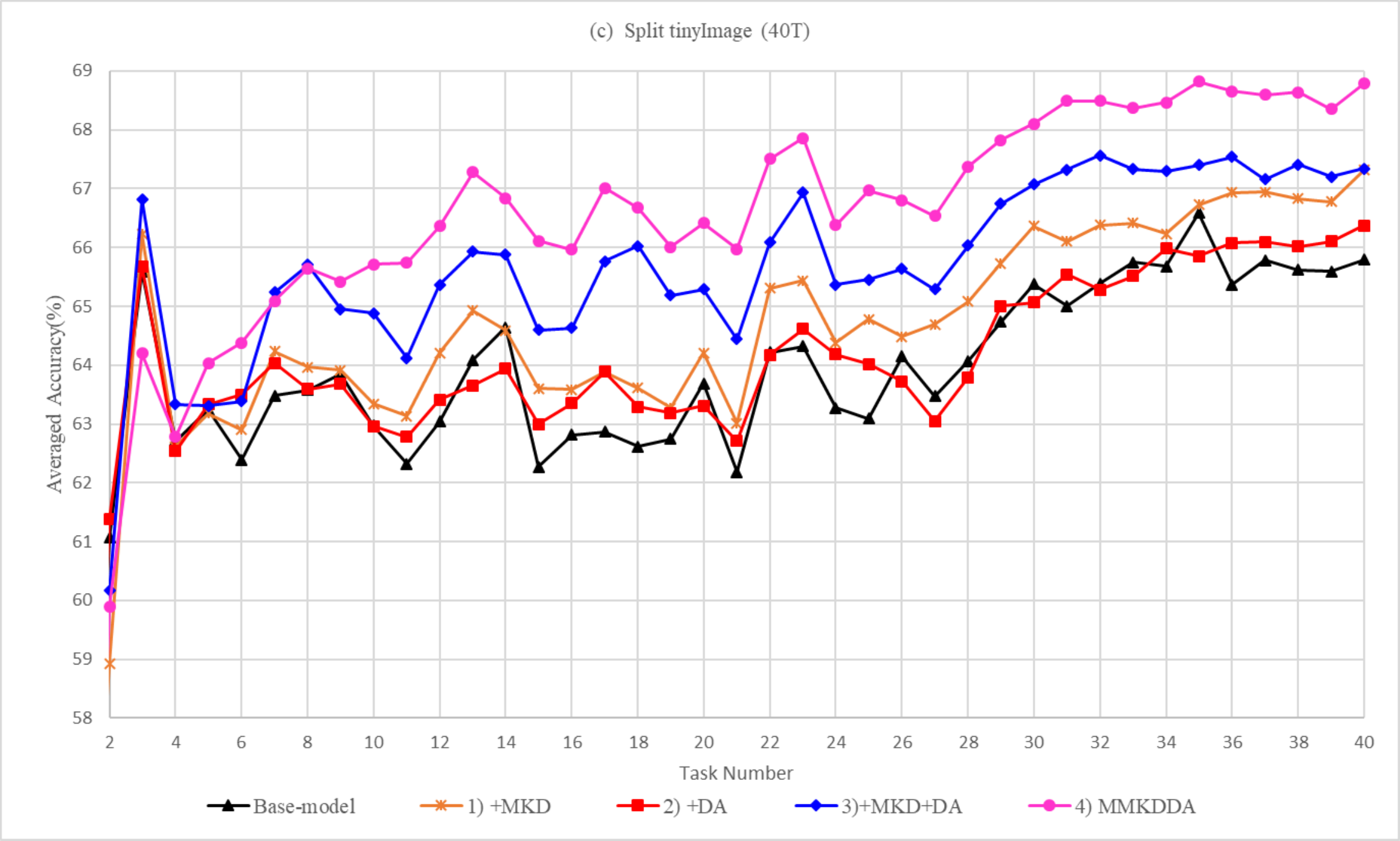}
	}
     \caption{Evolution of average accuracy as new tasks are learned}
     \label{fig9}
\end{figure}

\begin{figure}[H]
	\centering
	\subfigure[]{
		\includegraphics[width=0.8\textwidth]{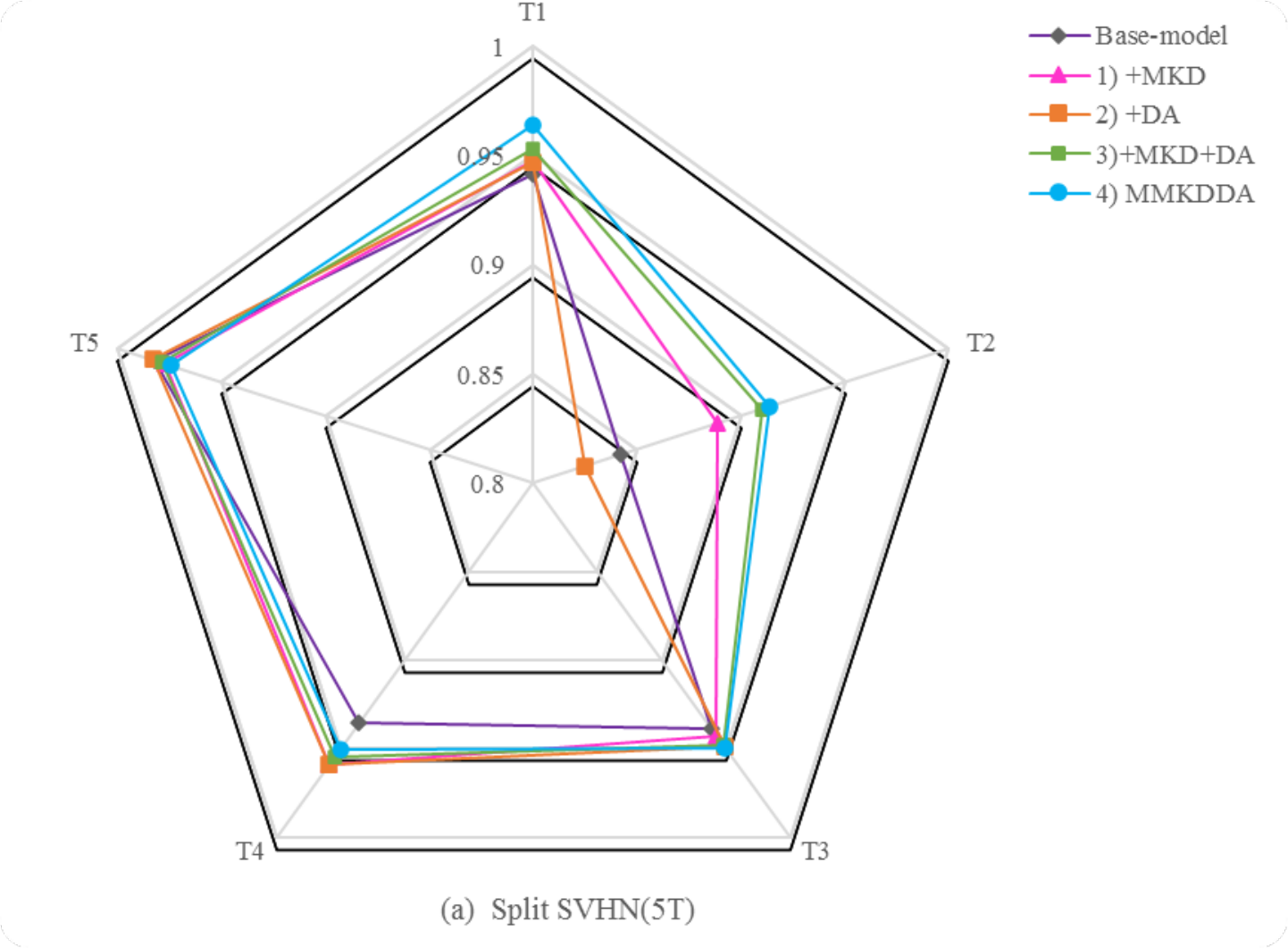}
	}
\end{figure}
\addtocounter{figure}{-1}
\begin{figure}[H]
\addtocounter{figure}{1}
	\centering
     \subfigure[]{
		\includegraphics[width=0.8\textwidth]{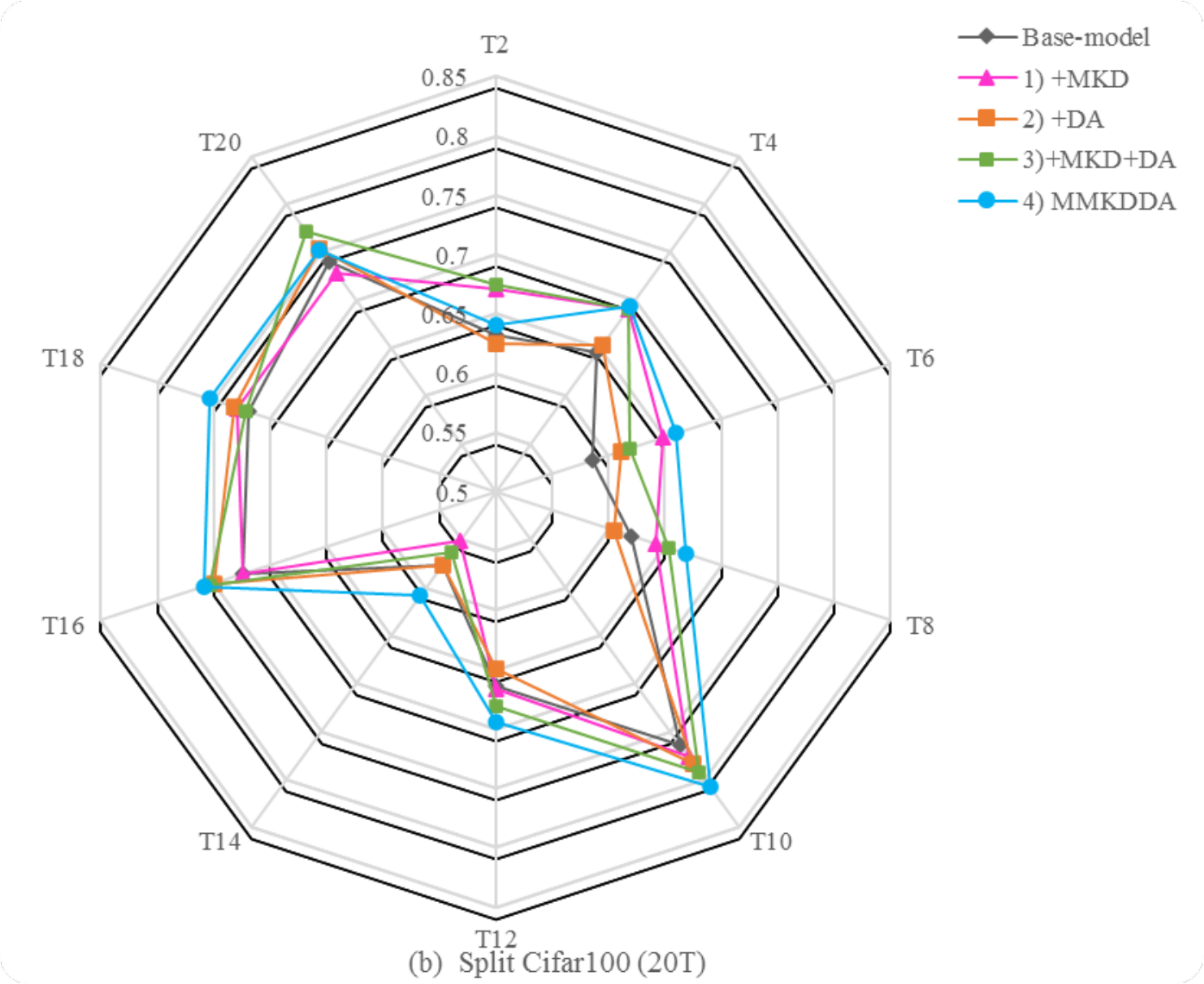}
	}
	\subfigure[]{
		\includegraphics[width=0.8\textwidth]{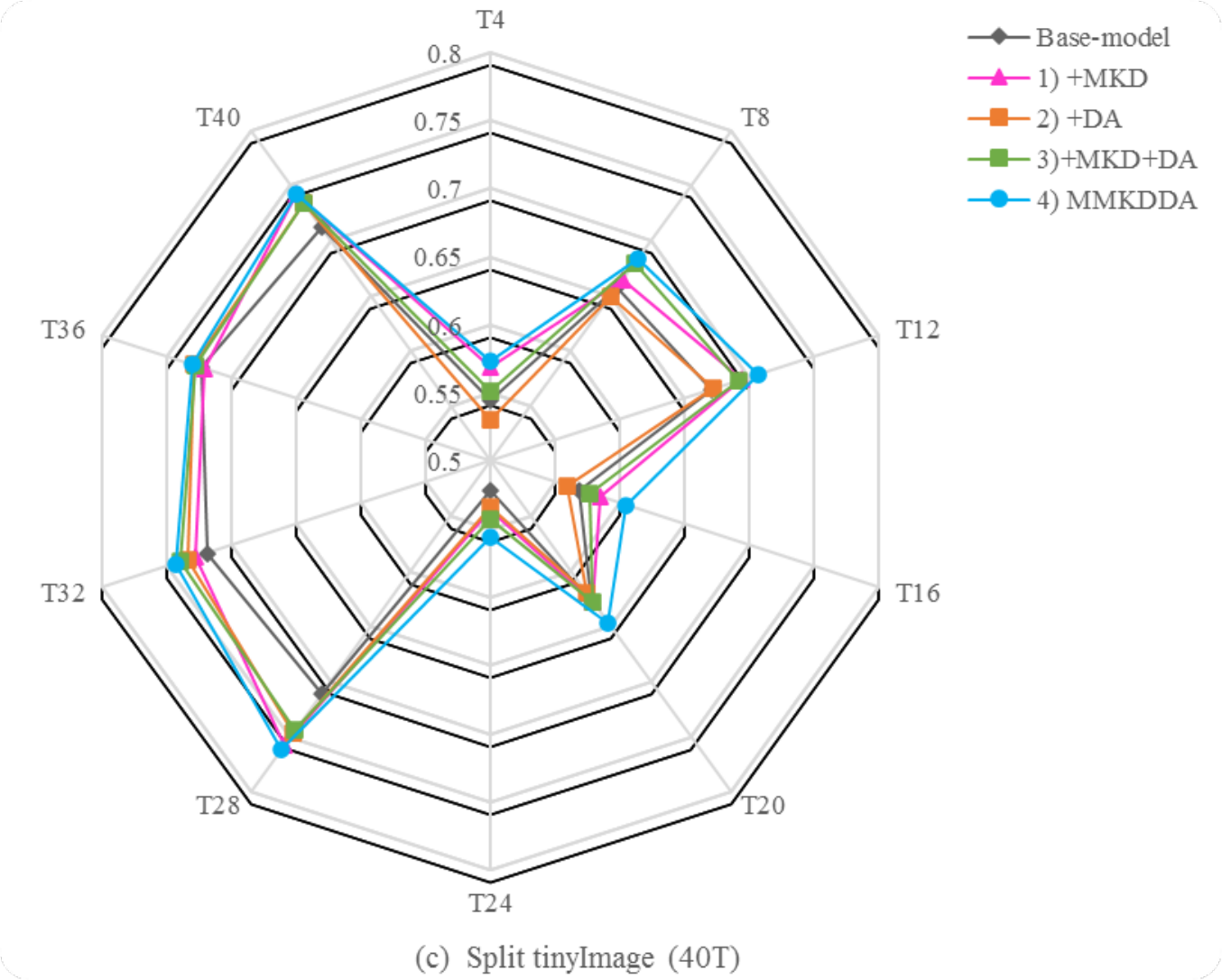}
	}
     \caption{The final accuracy at each task on different datasets}
     \label{fig10}
\end{figure}

\section{Conclusions and Future Works}

In this work, we studied the advantages and limitations of current continual learning approaches. Consider that although existing continual learning methods can obtain favorable results on some datasets, how to effectively keep stability while improving the ability to quickly learn new task remains a challenge and making a reasonable balance between stability and plasticity is not fully solved, especially in the context of online learning. Inspired by this, we propose a new framework, called MMKDDA, for online continual learning via a multi-scale knowledge distillation module, data augmentation strategy, and adaptive meta-learning. Specifically, the multi-scale knowledge distillation module can effectively alleviate the forgetting due to the imbalance between the new data and old data by constraining the evolution of long-range and short-range spatial relationships at different feature levels, which can further improve the stability. Furthermore, we also apply a data augmentation strategy to improve the plasticity, which means, making the model adapt to the desired task quickly. Lastly, the meta-learning update based on the number of tasks seen previously is used to further improve the ability, which makes the model learn to learn by leveraging the obtained knowledge, and can further make a balance between plasticity and stability via grasping the inherent relationships between consecutive tasks. Note that existing works either favor improving the plasticity of the model or focus on catastrophic forgetting. However, it is important to alleviate catastrophic forgetting while improving the ability to adapt to new tasks, especially in online continual learning. These limitations motivated us to propose a new framework, called MMKDDA. The extensive experiments demonstrate consistent improvements across a range of classification datasets. Furthermore, ablation studies can further understand the contribution of each aspect in our method. Such as, for the split cifar100 dataset, our method offers 68.73\% (1.22) to 69.74\% (0.59) LA improvements with data augmentation. LA reflects the model’s ability about learning new knowledge. When the multi-scale knowledge distillation and meta-learning update based on the number of tasks seen previously are employed, our model consistently offers 0.5\% to 2\% the final average accuracy (ACC) improvements. 

There are some avenues for future study. For example, in this work, we have shown the effectiveness of our MMKDDA method in classification scenarios and we can consider experiments are performed outside of our learning scenes, such as continual semantic detection. Besides, we also would attempt to apply our algorithm to a more realistic scenario, such as class incremental setting and domain incremental learning. Furthermore, we consider that generate models have captured many researchers’ attention and achieved tremendous success in recent years. Based on this, we also consider applying a generative model to generate new examples that are similar to previous tasks instead of episodic memory to overcome forgetting. Finally, current approaches for continual learning focus on learning a generic data representation for each task, which is not appropriate for this setting. In the future, we also attempt to propose an effective feature selection strategy. We can adaptively extract the relevant features for each task by this strategy. After that, we can further investigate how to improve the performance of the model.

\bibliography{bibfile}
\end{document}